\documentclass[accepted]{uai2022} %

\usepackage[american]{babel}

\usepackage{natbib} %
\renewcommand{\bibsection}{\subsubsection*{References}}
\usepackage{mathtools} %
\usepackage{booktabs} %
\usepackage{tikz} %

\usepackage{bm}
\usepackage{dsfont}
\usepackage{graphicx}
\graphicspath{{./figs/}}
\usepackage{caption}
\usepackage{subcaption}
\usepackage{wrapfig}
\usepackage{xcolor}
\usepackage{hyperref}

\definecolor{navyb}{RGB}{0,0,128}
\definecolor{burgundy}{RGB}{150, 0, 32}
\hypersetup{
  colorlinks,
  citecolor=navyb,
  linkcolor=burgundy,
  urlcolor=navyb
  }

\usepackage{amsmath}
\usepackage{amssymb}
\usepackage{amsthm}
\usepackage[capitalize,noabbrev]{cleveref}

\usepackage{algorithm}
\usepackage{algorithmic}

\newcommand{\hi}{\text{hi}}
\newcommand{\low}{\text{low}}

\newcommand\blfootnote[1]{%
  \begingroup
  \renewcommand\thefootnote{}\footnote{#1}%
  \addtocounter{footnote}{-1}%
  \endgroup
}

\usepackage{authblk}
\makeatletter
\renewcommand\AB@affilsepx{~~ \protect\Affilfont}
\makeatother

\title{Temporal Abstractions-Augmented Temporally Contrastive Learning: \\An Alternative to the Laplacian in RL}

\author[1,2]{Akram~Erraqabi}
\author[4,5,6]{Marlos~C.~Machado}
\author[1,3]{Mingde~Zhao}
\author[8]{Sainbayar~Sukhbaatar}
\author[8]{\authorcr{Alessandro~Lazaric}}
\author[8]{Ludovic~Denoyer}
\author[1,2,7]{Yoshua~Bengio}
\affil[1]{%
    Mila
}\affil[2]{%
    Université de Montréal
}
\affil[3]{%
    McGill University
  }
\affil[4]{%
    Amii\protect\\
  }
\affil[5]{%
    University of Alberta
  }
\affil[6]{%
    CIFAR AI Chair
  }
\affil[7]{%
   CIFAR Fellow
  }
\affil[8]{%
    Meta AI
  }
 
\begin{document}
\maketitle

\begin{abstract}
In reinforcement learning, the graph Laplacian has proved to be a valuable tool in the task-agnostic setting, with applications ranging from skill discovery to reward shaping. Recently, learning the Laplacian representation has been framed as the optimization of a temporally-contrastive objective to overcome its computational limitations in large (or continuous) state spaces. However, this approach requires uniform access to all states in the state space, overlooking the exploration problem that emerges during the representation learning process.
In this work, we propose an alternative method that is able to recover, \emph{in a non-uniform-prior setting}, the expressiveness and the desired properties of the Laplacian representation. We do so by combining the representation learning with a skill-based covering policy, which provides a better training distribution to extend and refine the representation. We also show that a simple augmentation of the representation objective with the learned temporal abstractions improves dynamics-awareness and helps exploration.
We find that our method succeeds as an alternative to the Laplacian in the non-uniform setting and scales to challenging continuous control environments.
Finally, even if our method is not optimized for skill discovery, the learned skills can successfully solve difficult continuous navigation tasks with sparse rewards, where standard skill discovery approaches are no so effective.
\end{abstract}

\section{Introduction}
With\blfootnote{Correspondance to AE: <\href{mailto:<akram.erraqabi@mila.quebec>?Subject=TATC paper}{akram.erraqabi@mila.quebec}>} the advent of deep reinforcement learning~\citep{Mnih2015}, representation learning~\citep[c.f.][]{bengio2013} has become one of the main topics of interest in reinforcement learning (RL). In fact, learning in environments with rich observations and complex dynamics~\citep[e.g.,][]{Bellemare2020} has motivated recent works on learning representations, e.g. controllable or contingent features \citep{bengio2017indep,Choi2019} on top of which one can potentially learn latent models in the perspective of planning \citep{hafner2019learning,nasiriany2019planning,schrittwieser2020mastering} and control \citep{watter2015embed,banijamali2018robust,hafner2020dream}.

\begin{figure}[t]
    \centering
    \includegraphics[width=\columnwidth]{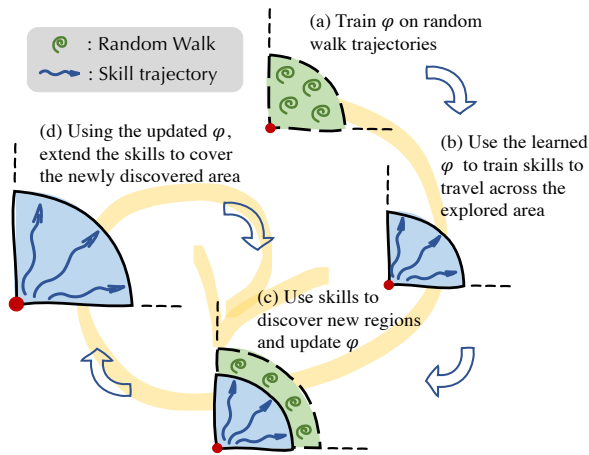}
    \caption{Our representation is trained to encode the area that the agent has learned to cover. Skills are continuously trained on the representation to discover new areas where novel data is collected to refine the representation, progressively extending its coverage. Similar incremental discovery is at the core of other works \citep{ecoffet2021nature, pong2019skew, machado2019thesis}.}
    \label{fig:algo_visual}
\end{figure}

In this work, we are interested in the task-agnostic setting in which an RL agent first interacts with the environment to build a representation, $\phi$, of the state space, $\mathcal{S}$, without relying on any specific reward signal. This representation can later be used to solve a task posed in the environment in the form of a reward function.
In this setting, the environment dynamics are the only informative interaction channel available to the agent. This has motivated the use of graph Laplacian-based methods where the graph vertices correspond to the states and its edges to the transitions probabilities. The Laplacian's eigenvectors can been leveraged as a holistic state representation, termed the Laplacian representation, which captures the environment dynamics~\citep{Mahadevan2005,mahadevan07aJMLR}.

\citet{Wu2019} have recently proposed an efficient approximation of the Laplacian representation (\textsc{Lap-rep}) by framing the graph drawing objective as a temporally-contrastive loss (see~\cref{sec:lap-rep-def}). This formulation works around potentially prohibitive eigendecompositions and extends the Laplacian's applicability to large (and continuous) state spaces. However, it assumes access to a uniform sampling prior over all states in the state space.
In practice, this translates in the ability to reset the agent to any state in the environment, which artificially alleviates the exploration problem.
As we show in~\cref{experiments}, this assumption is crucial for the quality of the learned representation. In the absence of the uniform prior privilege, such sampling is not trivial to achieve since the agent has to first explore and learn about the state space to be able to access arbitrary states. In effect, one must handle the exploration along the representation learning in order to preserve the quality of the representation. In this work, we propose \textsc{TATC}, an alternative representation learning framework to \textsc{Lap-rep} that extends a similar temporally-contrastive approach to a non-uniform-prior setting while preserving the desired properties and quality.

In practice, the representation is trained on data collected with a uniformly random policy. However, without a uniform access to the state space, the collected data is concentrated around accessible starting states. To achieve better data collection, we tie the representation learning problem to that of learning a covering strategy. Briefly, our method consists in using the available representation to train a skill-based covering policy that is in turn used to discover yet unseen parts of the state space, providing novel data to refine and expand the representation.
Our approach, illustrated in~\cref{fig:algo_visual} shares a similar motivation with several previous works~\citep{machado2019thesis, Machado2017,Machado2018,Jinnai2020Exploration}.
In addition to the aforementioned virtuous learning cycle between representation and skills, we propose to integrate the temporal abstractions captured by the skills in the contrastive representation learning objective. This augmentation contributes to a better temporally-extended exploration and enforce the representation's \emph{dynamics-awareness}, i.e. how representative the representation-induced metric is of distances in the state space.

We empirically show our agent's ability to progressively explore the state space and consistently extend the domain covered by representation in a non-uniform-prior setting. We show that our representation leads to better value predictions than \textsc{Lap-rep}, and that it recovers the representation quality expected from a uniform prior.
We also evaluate our representation in shaping rewards for goal-achieving tasks, and we show it outperforms \textsc{Lap-rep}, confirming both its superior ability in capturing dynamics and in scaling to \emph{larger} environments. Finally, the skills learned with our framework also prove to be successful at difficult continuous navigation tasks with sparse rewards, where other standard skill discovery methods have limited efficacy.

\section{Preliminaries}
\label{sec:background}

\subsection{Task-agnostic RL}

We describe a task-agnostic reinforcement learning (RL) environment as a task-agnostic Markov decision process (MDP) $\mathcal{M} = (\mathcal{S}, \mathcal{A}, P, \gamma, d_0)$ where $\mathcal{S}$ is the state space, $\mathcal{A}$ the action space, $P:\mathcal{S}\times\mathcal{A} \rightarrow{\Delta(\mathcal{S})}$ is the transition dynamics defining the next state distribution given the current state and action taken, $\gamma \in [0, 1)$ the discount factor, and $d_0$ is the initial state distribution. A policy $\pi: \mathcal{S}\rightarrow \Delta(\mathcal{A})$ maps states $s \in \mathcal{S}$ to distributions over actions. We denote by $\Delta$ the probability simplex.

Knowledge acquired from task-agnostic interactions with the environment (e.g., a representation or a policy) can then be leveraged for specific tasks.
A task is instantiated with a reward function, $R: \mathcal{S} \rightarrow \mathbb{R}$, which is combined with the task-agnostic MDP. The task objective is to find the optimal policy maximizing
the expected discounted return, $\mathbb{E}_{\pi, d_0} \Big[\sum_t \gamma^t R(s_t, a_t)\Big]$, starting from state $s_0 \sim d_0$ and acting according to $a_t \sim \pi(\cdot |s_t)$.

\subsection{The Laplacian Representation}
\label{sec:lap-rep-def}

The Laplacian representation (\textsc{Lap-rep}), as proposed by \citet{Wu2019}, can be learned with the following contrastive objective:
\begin{multline}
\label{eq:base_obj_general}
\mathcal{L}_{Lap}(\phi; \mathcal{D}_{\pi_\mu}) = \mathbb{E}_{(u, v)\sim \mathcal{D}_{\pi_\mu}} \Big[ \|\phi(u)-\phi(v)\|^2_2  \Big] + \\ \beta ~~ \mathbb{E}_{\substack{u\sim \mathcal{D}_{\pi_\mu}\\v \sim \mathcal{D}_{\pi_\mu}}} \Big[ (\phi(u)^\top\phi(v))^2-\|\phi(u)\|^2_2 -\|\phi(v)\|^2_2 \Big],
\end{multline}
where $\beta$ is a hyperparameter, $\phi:\mathcal{S}\rightarrow \mathbb{R}^d$ is a $d$-dimensional representation, $\pi_\mu$ the uniformly random policy (random walk trajectories), $\mathcal{D}_{\pi_\mu}$ a set of trajectories from $\pi_\mu$ (random walks). We use $(u, v) \sim \mathcal{D}_{\pi_\mu}$ to denote the sampling of a random transition from $\mathcal{D}_{\pi_\mu}$, and similarly $u \sim \mathcal{D}_{\pi_\mu}$ for a random state. \citet{Wu2019} showed the competitiveness of the Laplacian representation when provided with a uniform prior over $\mathcal{S}$ during the collection of $\mathcal{D}_{\pi_\mu}$. This objective is a \emph{temporally-contrastive} loss: it is comprised of an attractive term that forces temporally close states to have similar representations, and of a repulsive term that keeps far apart temporally far states' representations. Here, the repulsive term was specifically derived from the orthonormality constraint of the Laplacian eigenvectors.

\subsection{The non-uniform prior setting}
\label{sec:non_unif_prior}
In RL, representation learning is deeply coupled to the problem of exploration. Indeed, the induced state distribution defines the representation's training distribution. For instance, \textsc{Lap-rep} \citep{Wu2019} has been learned in the specific \emph{uniform prior}.
In this setting, $\mathcal{D}_{\pi_\mu}$, from ~\cref{eq:base_obj_general}, is a collection of trajectories with uniformly random starting states, which provides a uniform training distribution to the representation learning objective.
In the case of a non-uniform prior, the induced visitation distribution can be quite concentrated around the starting states distribution when solely relying on random walks to collect data, hence the need for a better exploration strategy in order to achieve a better training distribution for $\phi$.

To study the problem described above, we investigate the setting in which the environment has a fixed predefined state $s_0$ to which it resets with a probability $p_r$ every $K$ steps; with $K$ of the order of diameter of $\mathcal{S}$. With a uniformly random behavior policy, this setting is equivalent to a initial state distribution that is concentrated around $s_0$ and whose density decays exponentially away from it. We will refer to this setting as the \emph{non-uniform-prior} (non-$\mu$) setting, as opposed to the \emph{uniform-prior} ($\mu$) setting where the agent has uniformly access to the state space.

\section{Temporal abstractions augmented representation learning}
\label{method}

In this section, we present \underline{T}emporal \underline{A}bstractions-augmented \underline{T}emporally-\underline{C}ontrastive learning (\textsc{TATC}), a representation learning approach in which the representation works in tandem with a skill-based covering policy for a better representation learning in the non-uniform prior setting.

Before presenting the components of  \textsc{TATC} and how they are trained, we first provide an intuitive description of how it operates in the non-uniform prior setting. A detailed description of the algorithm can be found in \cref{app:algo}.

\subsection{TATC: a sketch of the algorithm}

In the non-uniform-prior setting, the agent is reset to a fixed state $s_0$ after each episode with some probability $p_r$. At the beginning of each episode (not necessarily at $s_0$, due to the probabilistic resetting), our agent can choose (with some probability $p_{rw}$) either to follow the uniform policy $\pi_\mu$ or to act according to an exploratory skill-based policy. Random walk data are, similarly to \citet{Wu2019}, collected to train the representation, while the skills are trained to extend the area of $\mathcal{S}$ that the agent is able to efficiently reach. In order to leverage the rich compositionality of skills, the agent executes a sequence of $L$ consecutive skills each time it decides to call the skill-based policy.
Initially, trajectories from $\pi_\mu$ cover the vicinity of $s_0$, making the representation reliable in that area, i.e. representative of its dynamics. Inevitably, the skills trained on this representation benefit from its emerging structure and progressively gain \emph{behavioral} structure: they allow the agent to travel efficiently across this explored area. In other words, the agent becomes capable of reaching the frontier of the explored regions faster, and it is able to collect,
using  $\pi_\mu$, novel data for the representation. The latter is hence refined, and its coverage extended. With a refined representation, the skills are able to reach even further areas.
This process emerges as a virtuous collaboration between the representation and the skills, where both benefit from improving each other. Eventually, by acquiring more knowledge from unexplored areas, the agent helps overcoming the loss in the representation's expressiveness that we observed when solely relying on $\pi_\mu$ in the non-uniform-prior setting.

In the remainder of this section we first propose a generic alternative objective to~\cref{eq:base_obj_general} that suits the non-uniform prior setting, then we describe the exploratory policy training. Finally, we introduce an augmentation of the proposed objective based on the learned temporal abstractions, to improve exploration and enforce the dynamics-awareness of the representation.

\subsection{Temporally-Contrastive Representation Objective}

As mentioned in Section~\ref{sec:lap-rep-def}, the repulsive term in \textsc{Lap-rep}'s objective, in \cref{eq:base_obj_general}, derives from the eigenvectors' orthonormality constraint.
However, because the environment is expected to be progressively covered in the non-uniform prior setting, the orthonormality constraint can make online representation learning highly non-stationary.\footnote{In general, even within a given matrix's perturbation neighborhood, its eigenvectors can show a highly nonlinear sensitivity~\citep{trefethen1997numerical}.}
For this reason, we adopt the following objective with a generic repulsive term that is more amenable to online learning:
\begin{multline}
\label{eq:base_obj_empir}
\mathcal{L}_{cont}(\phi; \mathcal{D}_{\pi_\mu}) \triangleq \mathbb{E}_{(u,v) \sim \mathcal{D}_{\pi_\mu}} \left[ \|\phi(u)-\phi(v)\|^2_2 \right] + \\ \beta ~~ \mathbb{E}_{\substack{u\sim \mathcal{D}_{\pi_\mu} \\ v \sim \mathcal{D}_{\pi_\mu}}}
\left[ \exp(-\|\phi(u)- \phi(v)\|_2) \right].
\end{multline}

In the following, we describe our representation-based skills training framework. Beyond addressing the exploration need, these skills will later (\cref{section:aug_rep_L}) be used to augment the objective above to obtain \textsc{TATC}'s representation learning objective.

\subsection{Representation-based Skills Training}
\label{section:hierarch_policy_train}

In the non-uniform-prior setting, exploration is required to provide the representation with a better training distribution. To this purpose, we adopt a hierarchical RL approach to leverage the exploration efficiency of skills, also known as options ~\citep{Sutton1999}. Let $\phi: \mathcal{S} \rightarrow \mathbb{R}^d$ be our $d$-dimensional representation. The agent acts according to a bi-level policy $(\pi_\hi,\pi_\low)$. The high-level policy $\pi_\hi: \mathcal{S} \to \Delta(\Omega)$ defines, at each state $s$, a distribution over a set $\Omega$ of directions (unit vectors) in the representation space ($\Omega=\{\bm{\delta} ~|~\bm{\delta} \in \mathbb{R}^d,  \|\bm{\delta}\|_2{=}1\}$). Each direction corresponds to a fixed length skill encoded by the low-level policy $\pi_\low: \mathcal{S}\times\Omega \rightarrow \Delta(\mathcal{A})$. These skills are trained to travel \emph{in the representation space} along the directions instructed by $\pi_\hi$.
In short, given a sampled direction $\pi_\hi(\cdot|s) \sim \bm{\delta} \in \Omega$, the low-level policy executes the skill $\pi_\low(\cdot|s, \bm{\delta})$ for a fixed number of steps $c$ before $\pi_\hi$ is called again.

Now, we describe the intrinsic rewards used to train the policies $\pi_\low$ and $\pi_\hi$.
\vspace{-1ex}
\paragraph{Low-level Policy.}
$\pi_\low$ is trained to follow directions picked by $\pi_\hi$ in the representation space. For a given $\bm{\delta} \in \Omega \subset \mathbb{R}^d$, the corresponding skill $\pi_\low(\cdot|s,\bm{\delta})$ is trained to maximize the reward function: 
\begin{equation}
\label{eq:skill_reward}
    r^{\bm{\delta}}(s,s') \triangleq \frac{(\phi(s') - \phi(s))^\top \bm{\delta}}{\|\phi(s') - \phi(s)\|_2}~~~,
\end{equation}
where $(s,s')$ is an observed state transition, and $\phi$ the representation being learned.
We use the cosine similarity as a way to encourage learning diverse directional skills. Indeed, skills co-specialization is avoided by rewarding the agent for the steps induced along the instructed direction $\bm{\delta}$ regardless of their magnitudes.
\vspace{-1ex}
\paragraph{Connection to Behavioral Mutual Information.} It is worth noting that our reward design can be interpreted as a mutual-information-based intrinsic control. We provide more details on this connection in \cref{app:connect_visr}.
\vspace{-1ex}
\paragraph{High-level Policy.}
The high-level policy guides the covering strategy. It does so by sampling the skills of the most promising directions in terms of the amount of exploration, measured by the travelled distance in the representation space.
For this purpose, we design a reward function defined over a sequence of $L$ consecutive skills. Let $\{s^\hi_k\}_{k=1}^L$ be the sequence of their initial states and their respective sampled directions, $\bm{\delta}_k \sim \pi_\hi(\cdot|s^\hi_k)$. Since $\phi$ is trained to capture the dynamics, the travelled distance in $\phi$'s space \emph{expresses} how far the choices made by $\pi_\hi$ eventually brought the agent in the environment.
Therefore, for a given high-level trajectory, $\tau^\hi=(s^\hi_1, s^\hi_2, ..., s^\hi_L,s^\hi_f)$, with $s^\hi_f$ the final state reached by the last skill, the high-level policy is trained to maximize the following:
\begin{equation}
\label{eq:high_reward}
   \forall k\in \{1,...,L\}, R^\hi(s^\hi_k, \bm{\delta}_k) \triangleq \|\phi(s^\hi_1) - \phi(s^\hi_f) \|_2 ~~,
\end{equation}
where $\bm{\delta}_k \sim \pi_\hi(\cdot|s^\hi_k)$ is the direction sampled at $s^\hi_k$. From the policy optimization perspective, each of these quantities plays the role of the return cumulated along the sampled high-level trajectory and \emph{not} just a single (high-level) step reward. This term looks at reaching $s^\hi_f$ as the result of a sequential collaboration of $L$ skills, rewarding them equally. It values how far this sequence of skills eventually brought the agent.

In the following section, we show how the skills trajectories can be used to augment the representation with the learned temporal abstractions.
\subsection{Augmenting Representation Learning with Temporal Abstractions}
\label{section:aug_rep_L}
A skill abstracts a temporally-extended behavior in a single (high-level) action. As a \emph{temporal abstraction}, it represents a factorized knowledge of the environment dynamics in the form of a policy. Here, we propose to integrate these temporal abstractions to the representation objective to better capture the environment dynamics. In order to preserve the temporal contrast of the base objective (\ref{eq:base_obj_empir}), we augment it with the following contracting term along skills trajectories:
\vspace{-1ex}
\begin{equation}
\label{eq:boredom}
\mathcal{B}(\phi; \mathcal{D}_s) \triangleq ~~~~~~ \mathop{\mathbb{E}}_{\mathclap{\substack{ \tau_{\bm{\delta}} \sim \mathcal{D}_s \\ 
\tau_{\bm{\delta}} = (s_0, ..., s_c)
}}} ~~~~~~~~  \left[ \sum_{k=0}^{c-1} \| \phi(s_k) - \phi(s_{k+1}) \|_2 \right],
\end{equation}
where $\mathcal{D}_s$ is a set of collected skills trajectories.
By minimizing this term, $\phi$ integrates temporally-extended dynamics: areas connected by skills are brought closer in the representation space.
This term will be referred to as the \emph{boredom} term. Its exploratory virtue is discussed in the following.

\paragraph{How does boredom help exploration ?} The interplay between the high-level policy reward function (\ref{eq:high_reward}) and this boredom term (\ref{eq:boredom}) induces a progressive exploration mechanism. In effect, $\pi_\hi$ tends to sample skills that travel further, i.e. with larger $R^\hi$. The more often a skill is sampled, the less rewarding it becomes to $\pi_\hi$ due to the minimization of $\mathcal{B}(\phi)$. This will favor sampling the remaining (under-sampled) skills, hence encouraging the exploration of less visited parts of the state space.
This mechanism dynamically fights what can be considered as accumulated \emph{boredom} along over-sampled skills trajectories which increases the agent curiosity and urge it to explore.

Finally, the proposed objective to train the representation $\phi$ consists in the base objective (\ref{eq:base_obj_empir} augmented with the boredom term (\ref{eq:boredom}), and can be written as
\begin{equation}
\label{eq:rep-L}
\mathcal{L}_\text{TATC}(\phi; \mathcal{D}_s, \mathcal{D}_{\pi_\mu}) \triangleq \mathcal{L}_{cont}(\phi; \mathcal{D}_{\pi_\mu}) + \beta' \mathcal{B}(\phi; \mathcal{D}_s) ~,
\end{equation}
with $\beta'$ a hyperparameter controlling the strength of boredom term.

\section{Experiments}
\label{experiments}
In this section, we investigate the behavior of \textsc{TATC} in two types of environments: gridworld environments with discrete state and action spaces, and continuous navigation environments for continuous state spaces (MuJoCo,~\citet{todorov2012mujoco}).

For our method, we learn a 2D representation ($d=2$), and define $\Omega$ as a set of $8$ unit vectors equally spaced on the unit sphere. These directions are concatenated to the input state of $\pi_\low$. Implementation details of all the experiments in this section can be found in \cref{app:implem}.

\subsection{GridWorld}
\begin{figure}[H]
    \centering
    \begin{subfigure}[t]{0.275\columnwidth}
        \centering
        \includegraphics[width=\linewidth]{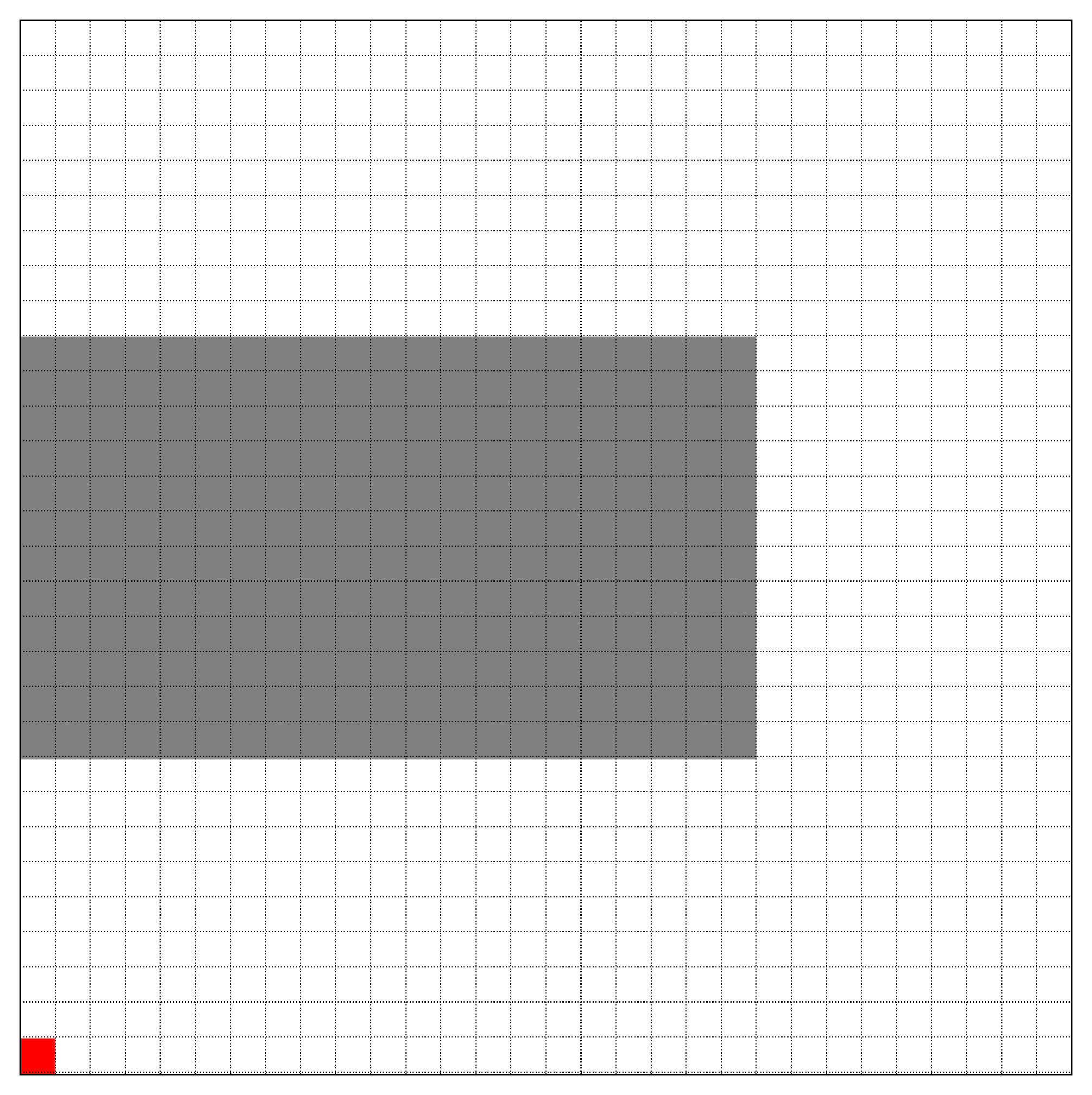}
        \caption{\textsc{U-Maze}}
    \end{subfigure}%
    \hspace{0.02\columnwidth}
    \begin{subfigure}[t]{0.36\columnwidth}
        \centering
        \includegraphics[width=\linewidth]{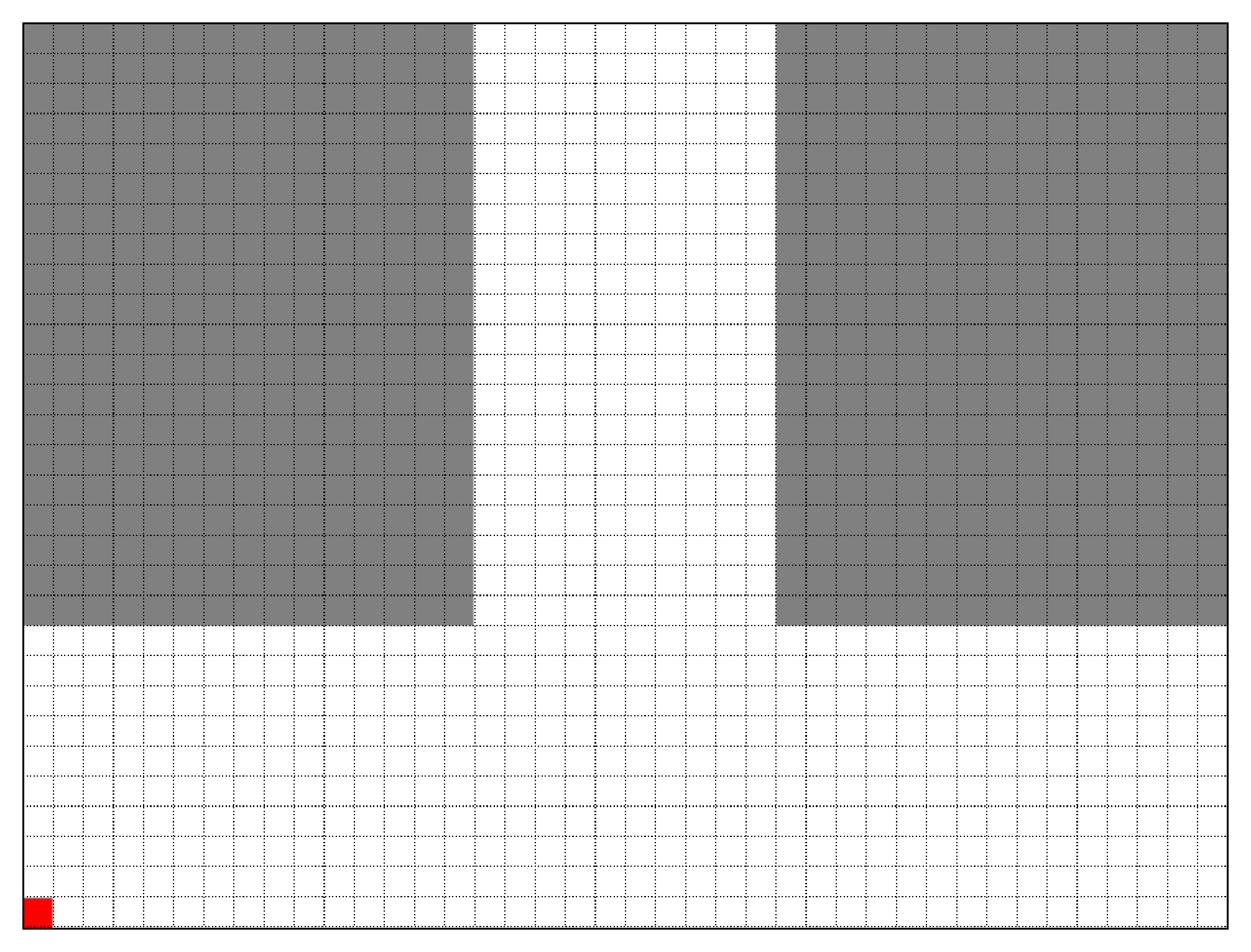}
        \caption{\textsc{T-maze}}
    \end{subfigure}
    \hspace{0.007\columnwidth}
    \begin{subfigure}[t]{0.27\columnwidth}
        \centering
        \includegraphics[width=\linewidth]{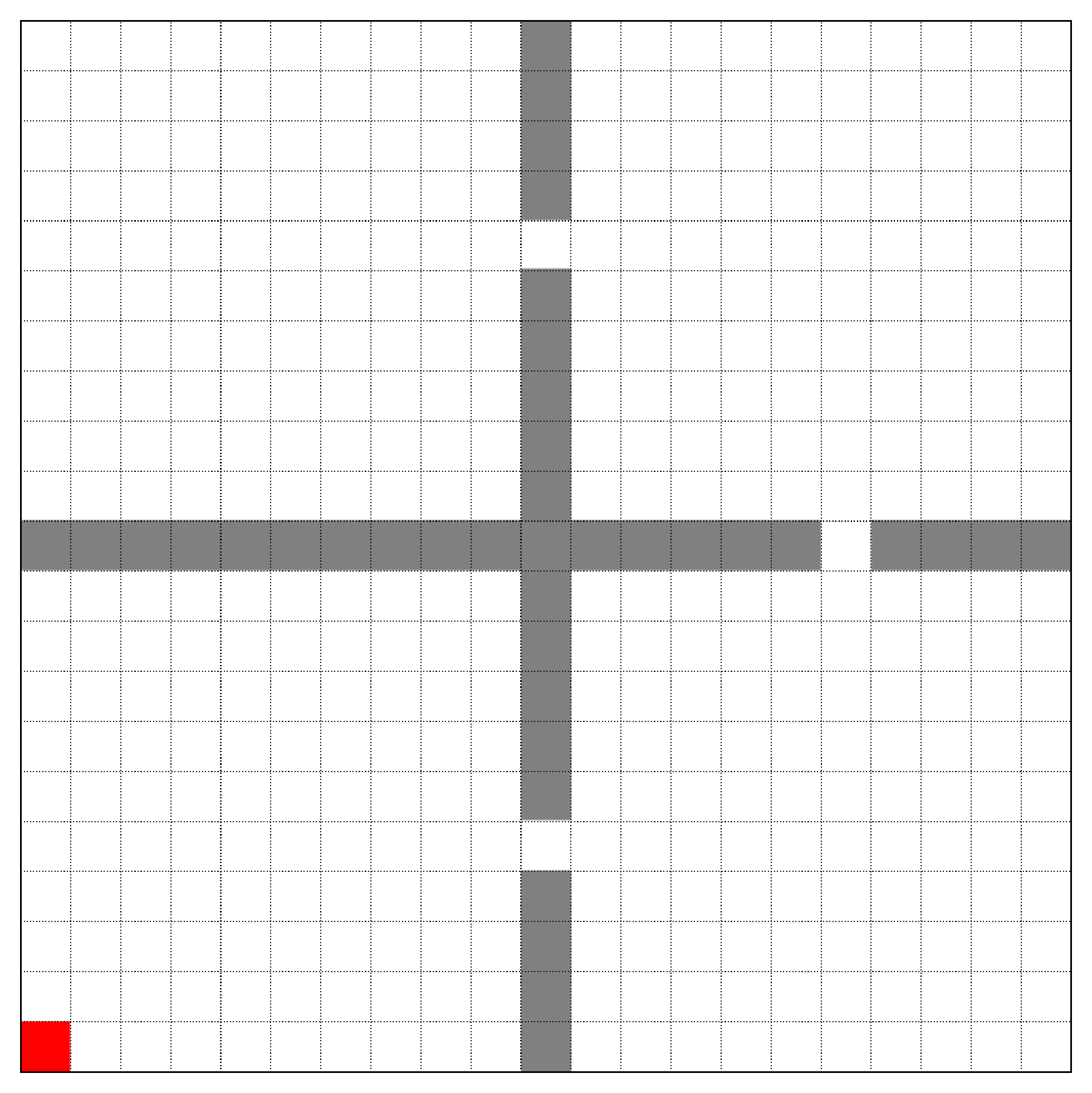}
        \caption{\textsc{4-rooms}}
    \end{subfigure}
\caption{Gridworld domains. The fixed initial state $s_0$ is highlighted in red.}
\label{fig:envs}
\end{figure}

We evaluate our approach in three gridworld domains: \textsc{U-Maze}, \textsc{T-Maze} and \textsc{4-rooms}. These environments, visualized in Figure \ref{fig:envs}, raise different explorations challenges. \textsc{U-Maze} is a simple but relevant environment to test the dynamics-awareness of the representations;\footnote{The presence of the wall makes L2-distance in xy-coordinates deceptive. The L2-distance in a dynamics-aware representation space should correct for that.} \textsc{T-Maze} raises the challenge of splitting the exploration focus at an intersection while maintaining the covering in both corridors; \textsc{4-rooms} is similar to \textsc{U-Maze}\footnote{Note that there is no door between the first and the fourth room.}, but requires learning more controlled skills for a useful exploration.

\begin{figure*}[t]
    \centering
    \includegraphics[width=\textwidth]{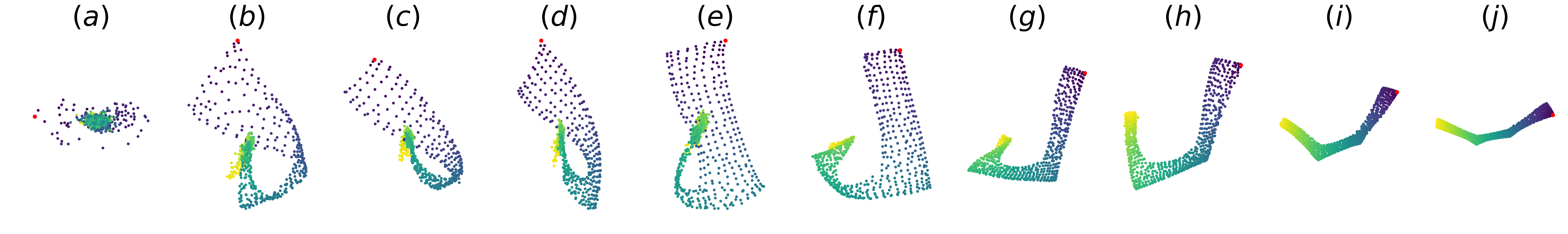}
    ~
    \includegraphics[width=\textwidth]{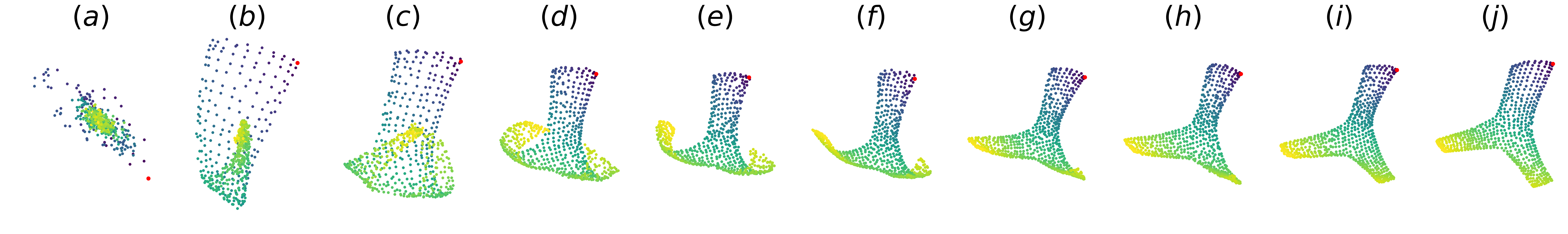}
    ~
    \includegraphics[width=\textwidth]{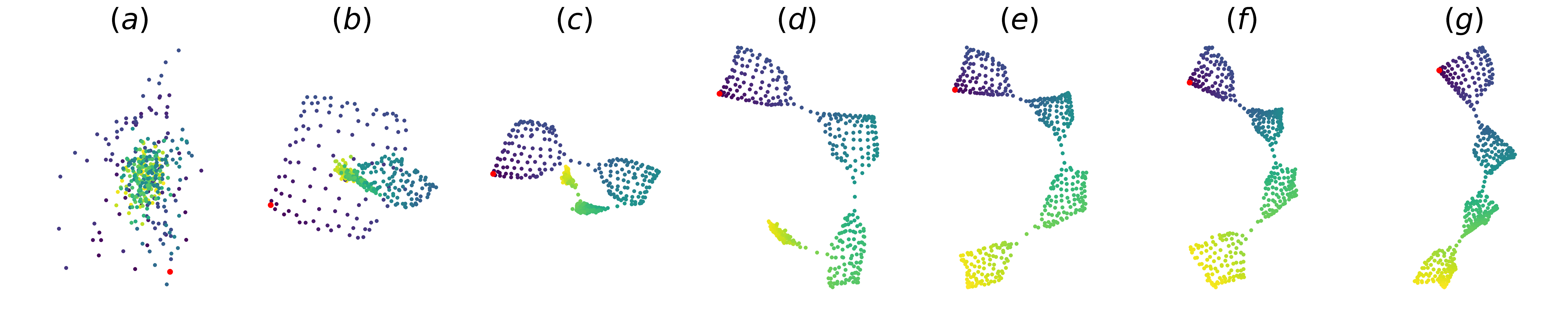}
    \caption{\textsc{TATC} representations learned throughout the training. For each domain, all the states are mapped with $\phi$ and represented at different stages of the training procedure. Axes scales were equalized for reliable visual appreciation. Top row: \textsc{U-Maze}. Middle row: \textsc{T-Maze}. Bottom row: \textsc{4-rooms}. The colors reflect the distances in terms of the dynamics. They can be seen as quantities proportional to the length of the shortest path from $s_{0}$ (marked in red) to the represented state.}
    \label{fig:progress_exp}
\end{figure*}

\vspace{-4ex}
\subsubsection{Progressive Representation Learning}

\cref{fig:progress_exp} shows the evolution of the representations throughout training. The agent progressively explores the environment starting around $s_0$, and builds the representation by continuously integrating newly discovered parts.

\textbf{\textsc{U-Maze} \& \textsc{4-rooms}}. The agent starts from the bottom left corner, or room, of the maze. \cref{fig:progress_exp} shows how the representation progressively expands away from $s_0$ along the corridor, or the rooms sequence. Note that while the agent learns to reach and represent further areas, the full domain representation \emph{flattens}, 
indicating the representation's success in capturing the maze dynamics.

\textbf{\textsc{T-Maze}}. The agent starts from the bottom left corner of the maze. As in the U-Maze, it starts learning to travel along the corridor until it reaches the intersection. There, the exploration focus is shared between both possible paths whose representations are progressively disentangled. Eventually, the agent fully explores both corridors and finalizes its representation. Note that, the discovery of one of the corridors did not hinder finishing the discovery of the other. The boredom term proved to be important for such property (see Appendix \ref{app:ablation}).

\textbf{Importance of Boredom.} Appendix~\ref{app:ablation} provides an ablation study showing the importance of the boredom term for the agent's exploratory behavior and the representation's dynamics-awareness.

\subsubsection{Evaluating the Learned Representation}
\label{sec:gridworld_quantitative}
We now compare our representation against \textsc{Lap-rep}~\citep{Wu2019} in the non-uniform-prior setting. First, to appreciate the sensitivity of \textsc{Lap-rep} to the uniformity of said prior, we trained \textsc{Lap-rep} in two settings: (i) the uniform-prior setting where the agent can be set to any arbitrary state, as done by~\citet{Wu2019}, and (ii) the non-uniform-prior setting defined in~\cref{sec:non_unif_prior}.
In the following, we show that \textsc{Lap-rep} is sensitive to this change in distribution while \textsc{TATC} recovers the expressive potential that a uniform prior provides.

\begin{figure*}[t]
\centering
\includegraphics[width=0.85\textwidth]{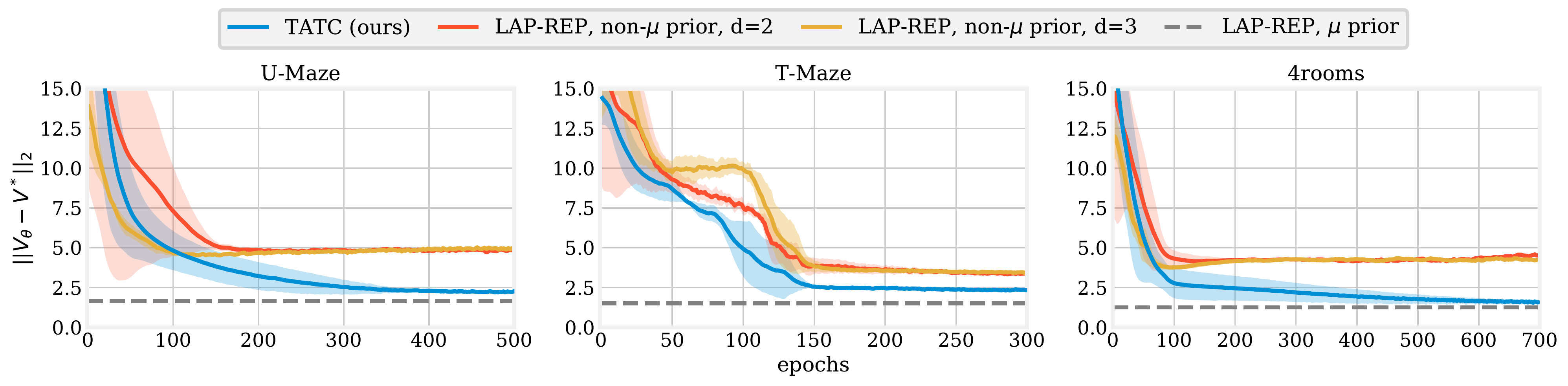}
\caption{Learned representation's ability to approximate the value function. \textsc{Lap-rep} was learned in the same non-uniform-prior setting (non-$\mu$) with $d=2$ and $d=3$ (no improvement was observed for higher values). The dashed line gives the performance of \textsc{Lap-rep} in the uniform-prior setting ($\mu$). \textsc{TATC} outperforms \textsc{Lap-rep} in non-$\mu$ setting, and succeeds in recovering its expressive power when learned from the uniform prior. Performances were averaged over 5 different runs. $95\%$ confidence intervals are shaded.}
\label{fig:LFA}
\end{figure*}

\textbf{Prediction.}
To evaluate the learned representations, we first consider how well they linearly approximate a given task's optimal value function. To do so, we train an actor-critic agent~\citep{mnih2016AC} with a linear critic on top of each representation. In Figure \ref{fig:LFA}, we note a significant loss in the representational power of \textsc{Lap-rep} when the access to the state space is not uniformly distributed anymore. This figure also shows that \textsc{TATC} outperforms \textsc{Lap-rep} in the non-uniform-prior setting, and succeeds in recovering \textsc{Lap-rep}'s expressive power when it is learned with the unrealistic uniform prior.

\begin{figure*}[t]
\centering
\includegraphics[width=0.8\textwidth]{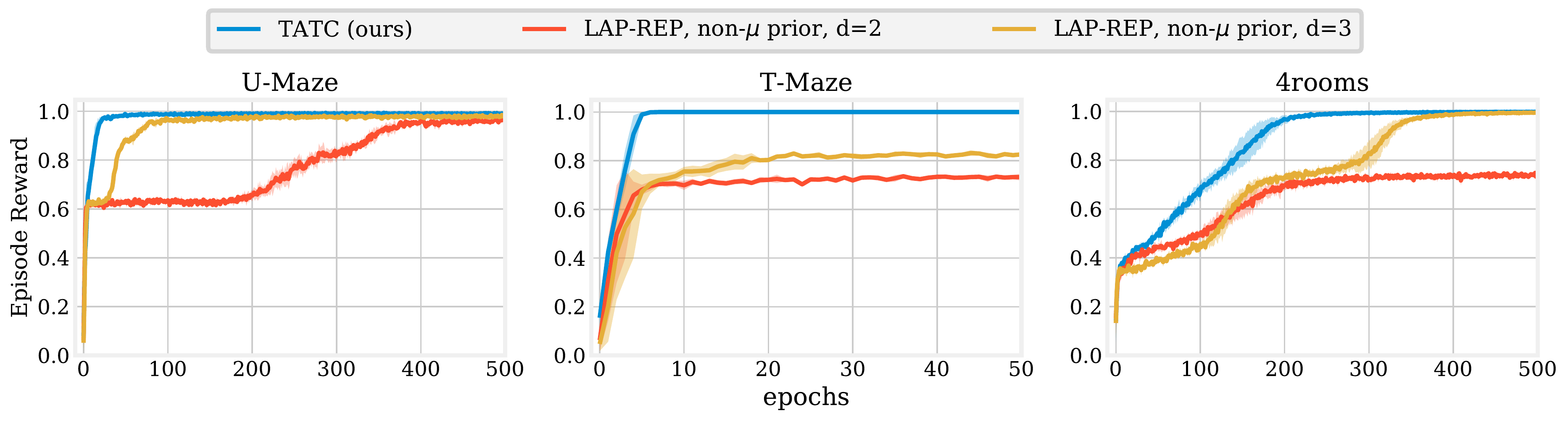}
\caption{Control performance (Episode Reward) in the non-uniform-prior setting. Performances were averaged over 5 different runs. $95\%$ confidence intervals are shaded.}
\label{fig:grid_control}
\end{figure*}

\textbf{Control.} We also compare the representations from the perspective of control, by training a deep actor-critic agent on top of each representation to solve a goal-reaching task in the same domains as above. The agent is only rewarded upon reaching the goal state ($r=1$). Figure~\ref{fig:grid_control} shows that \textsc{TATC} consistently outperforms \textsc{Lap-rep}, which confirms the competitive quality of our representation.

\subsection{Continuous Control}

\begin{figure}[H]
\centering
\begin{subfigure}[t]{0.48\columnwidth}
    \centering
    \includegraphics[width=\linewidth]{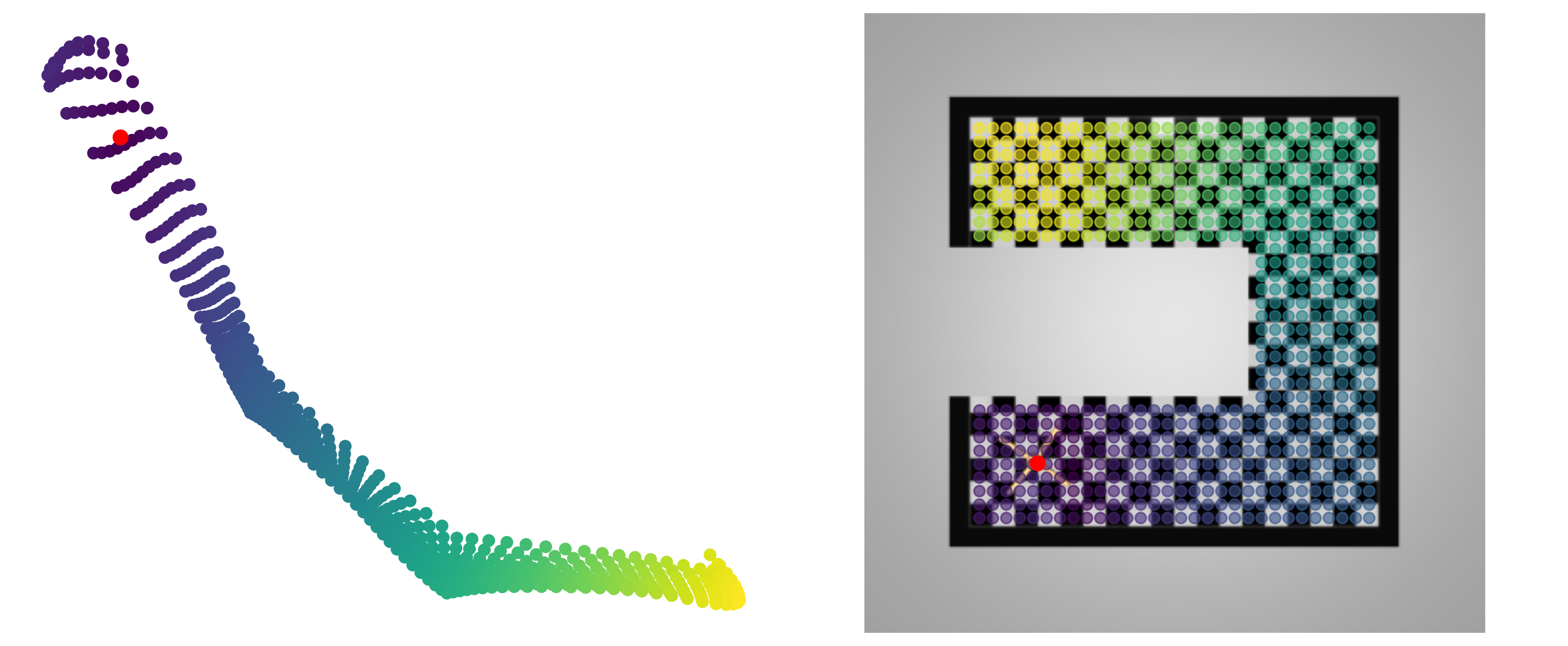}
    \caption{\textsc{AntMaze-1}}
\end{subfigure}
\begin{subfigure}[t]{0.49\columnwidth}
    \centering
    \includegraphics[width=\linewidth]{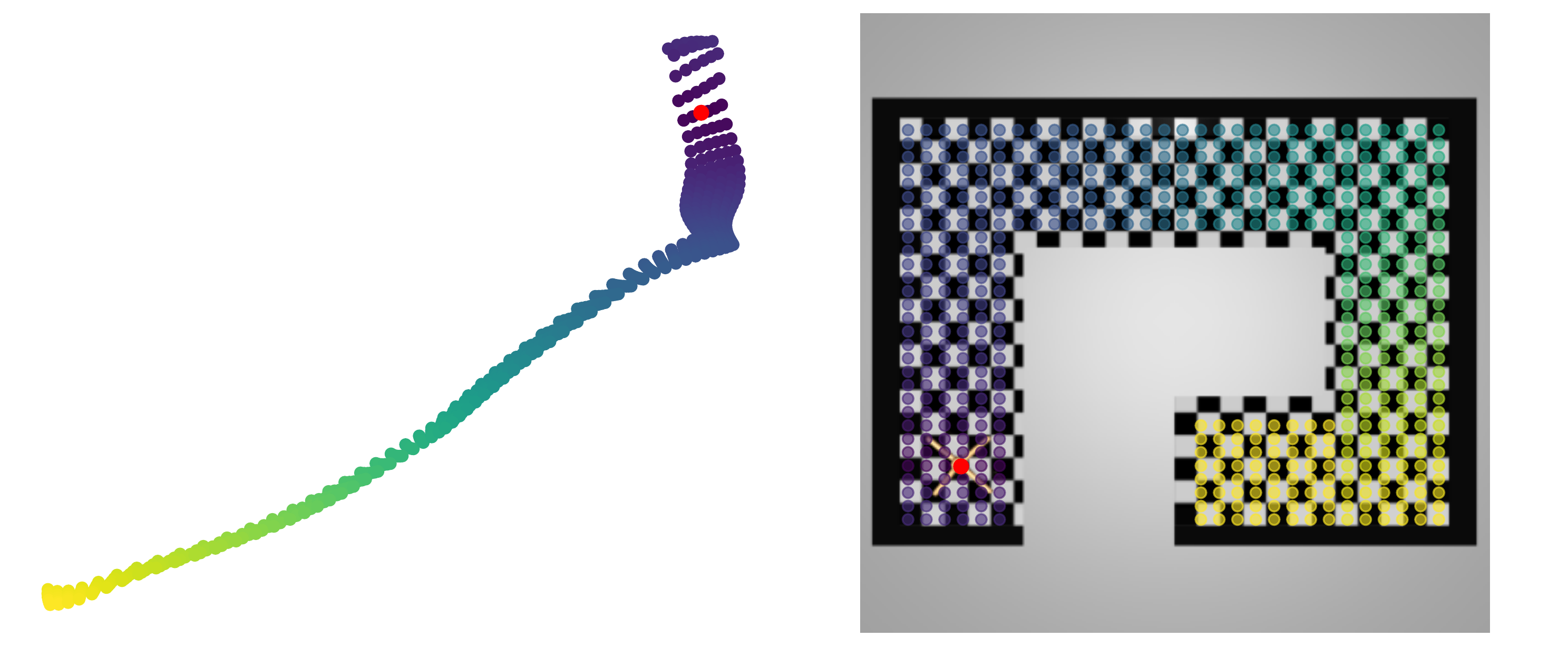}
    \caption{\textsc{AntMaze-2}}
\end{subfigure}%
\caption{Learned \textsc{TATC} representations visualized on grids of positional states. Colors reflect the distance in the representation space from the initial state, highlighted in red. Axes scales were equalized. We can visually appreciate how the continuous state domains are mapped to a flatter manifold reflecting the presence of the walls.}
\label{fig:antmaze_rep}
\end{figure}

The second set of experiments focuses on continuous state and action spaces. Here, we consider two mazes for the MuJuCo Ant agent, as shown in \cref{fig:antmaze_rep}. These are similar in shape to the ones from \citet{Wu2019}, but are larger and thus more challenging. More details in \cref{app:implem}.

To visualize our learned representations in these environments, \cref{fig:antmaze_rep} depicts a grid of positional states in each environment domain and their mapped representations. Similarly to \textsc{U-Maze} and \textsc{4-rooms}, the learned representation translates the environment dynamics, and specifically the presence of walls, by mapping the original state space to a flatter manifold.
\vspace{3ex}

\vspace{-3ex}

\subsubsection{Reward Shaping with Learned Representations} 
We demonstrate how \textsc{TATC} representation is able to improve an RL agent's performance when the distances in the representation space are used for reward shaping, the same setting in which \citet{Wu2019} evaluated \textsc{Lap-rep}.
We define a goal-achieving task by setting a goal state $g$ at the end of the corridor (visualized in \cref{app:implem}). The objective is to learn to navigate to a state $s$ close enough to the goal area ($\|s-g\|_2 \leq \epsilon$).
We define the reward function based on the distance in representation space (\textsc{TATC} and \textsc{Lap-rep}). More specifically, we train a soft actor-critic (SAC) agent~\citep{haarnoja2018soft} to reach the goal with a \textbf{dense} reward defined as $r^{dense}_t=-\|\phi(s_{t+1}) - \phi(g)\|_2$. Similarly to \citet{Wu2019}, we also compare against the half-half \textbf{mix} of the dense reward and the sparse reward $r^{mix}_t= 0.5\cdot r^{dense}_t + 0.5\cdot\mathds{1}\left [\|s_{t+1}-g\|_2\leq \epsilon \right ]$.

\begin{figure}[H]
    \centering
    \begin{subfigure}[t]{0.87\columnwidth}
        \centering
        \includegraphics[width=\linewidth]{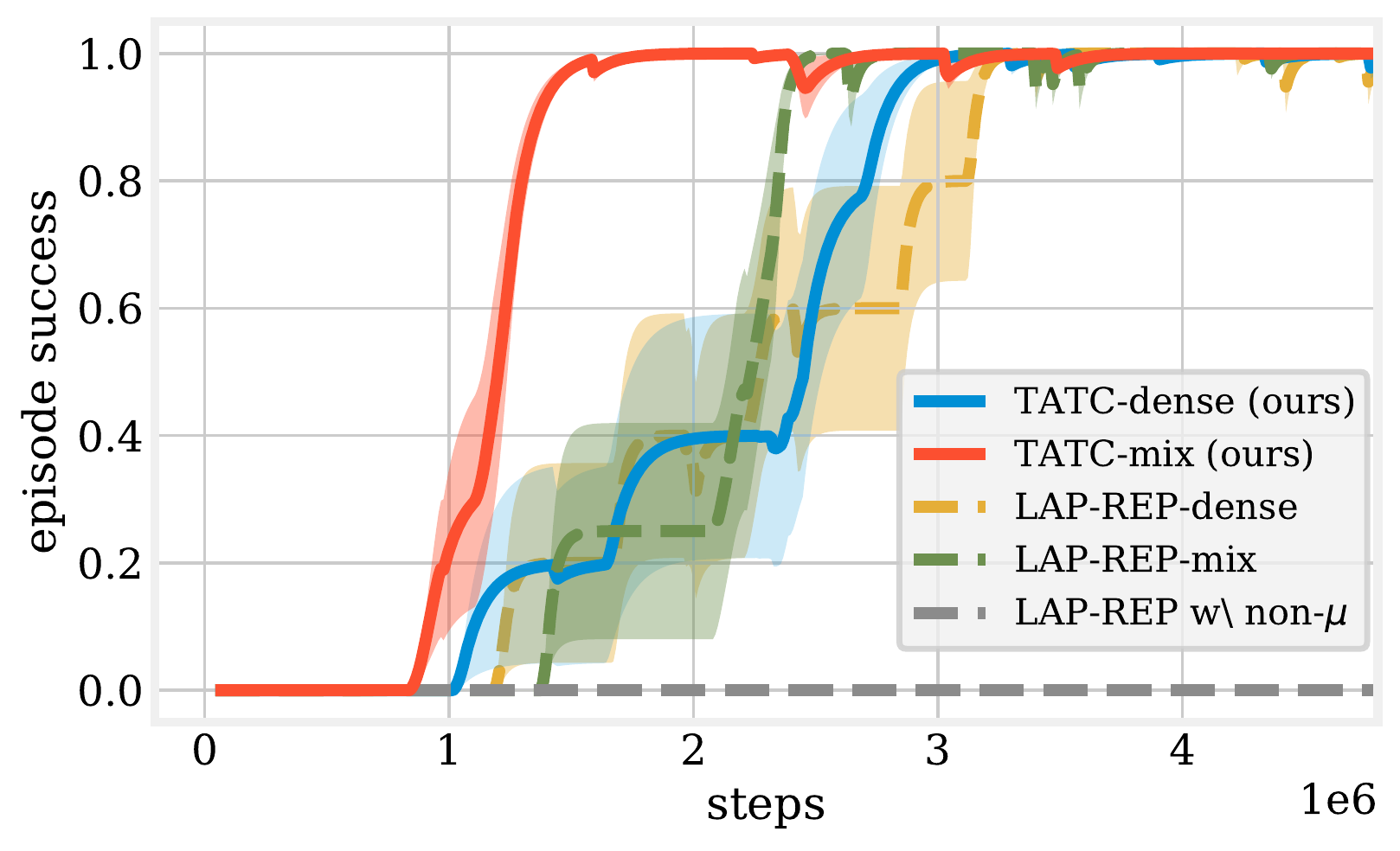}
        \caption{\textsc{AntMaze-1}}
    \end{subfigure}%
    \\
    \begin{subfigure}[t]{0.87\columnwidth}
        \centering
        \includegraphics[width=\columnwidth]{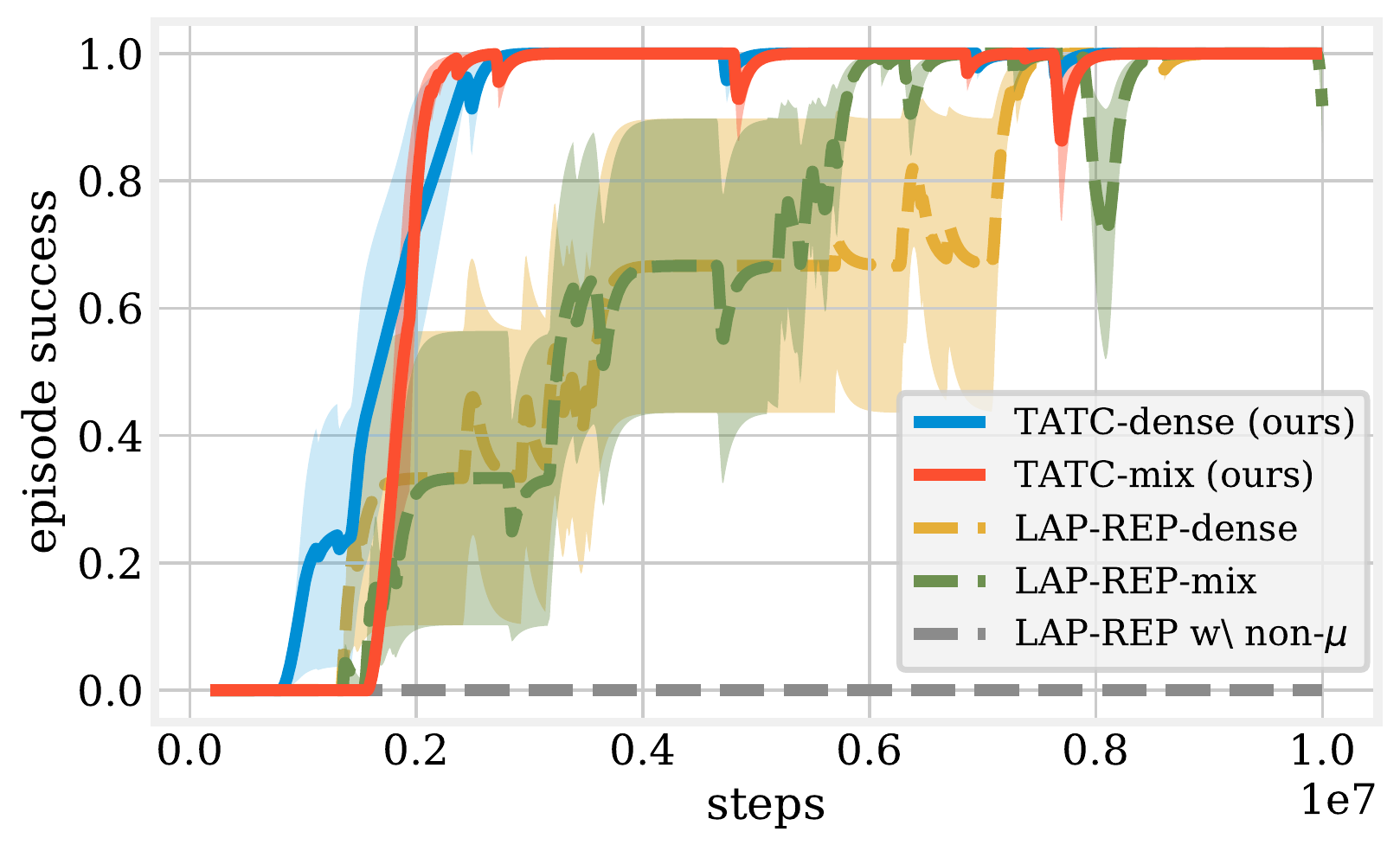}
        \caption{\textsc{AntMaze-2}}
    \end{subfigure}
\caption{Reward shaping using learned representations: performances were averaged over 5 different runs. $95\%$ confidence intervals are shaded. Curves were exponentially smoothed (0.9) for better visualization.}
\label{fig:rew_shp}
\end{figure}

For this evaluation, \textsc{Lap-rep} was learned, unlike our representation, with a \emph{uniform} prior over $\mathcal{S}$ as in \citet{Wu2019}, and $d=2$.\footnote{Our attempts with $d=20$ did not succeed at these reward shaping tasks.} Figure~\ref{fig:rew_shp} shows that our representation is effective in reward shaping, with both \textbf{mix} and \textbf{dense} variants
, and enjoys a comparable if not superior dynamics-awareness to \textsc{Lap-rep}. This result stands while \textsc{TATC} is learned from a non-uniform prior which is less advantageous than the uniform prior used to train \textsc{Lap-rep}. Note that \textsc{Lap-rep} with a non-uniform prior is unable to guide the agent to the goal.

Finally, these results further confirm the conclusions drawn from the gridworld experiments and positions \textsc{TATC} as a competitive alternative to \textsc{Lap-rep} in this challenging setting.

\subsubsection{Evaluating the Learned Skills}
To evaluate the exploratory potential of \textsc{TATC}'s skills, we compare the learned skills in \textsc{AntMaze-1} against 2 task-agnostic skill discovery methods, DIAYN \citep{Eysenbach19}, and DCO \citep{Jinnai2020Exploration}. DIAYN is a mutual information-based approach to learn diverse set of skills, while DCO skills are based on a temporally-contrastive representation, similarly to \textsc{TATC}. More specifically, DCO requires a pretrained \textsc{Lap-rep} which approximates the Laplacian's second eigenvector; also called the Fiedler vector. We train the required representation, as well as DCO, with the advantage of data collected from a \emph{uniform} prior over $\mathcal{S}$. For a fair comparison, we train $8$ skills for both methods (DCO and DIAYN).
Once trained, the skills learned by each method are fixed and used to train a discrete high-level policy that can select across the available skills to solve a goal-reaching task with a sparse reward function $r_t = \mathds{1}\left [\|s_{t+1}-g\|_2\leq \epsilon \right ]$. 

\begin{figure}[H]
    \centering
    \includegraphics[width=0.8\columnwidth]{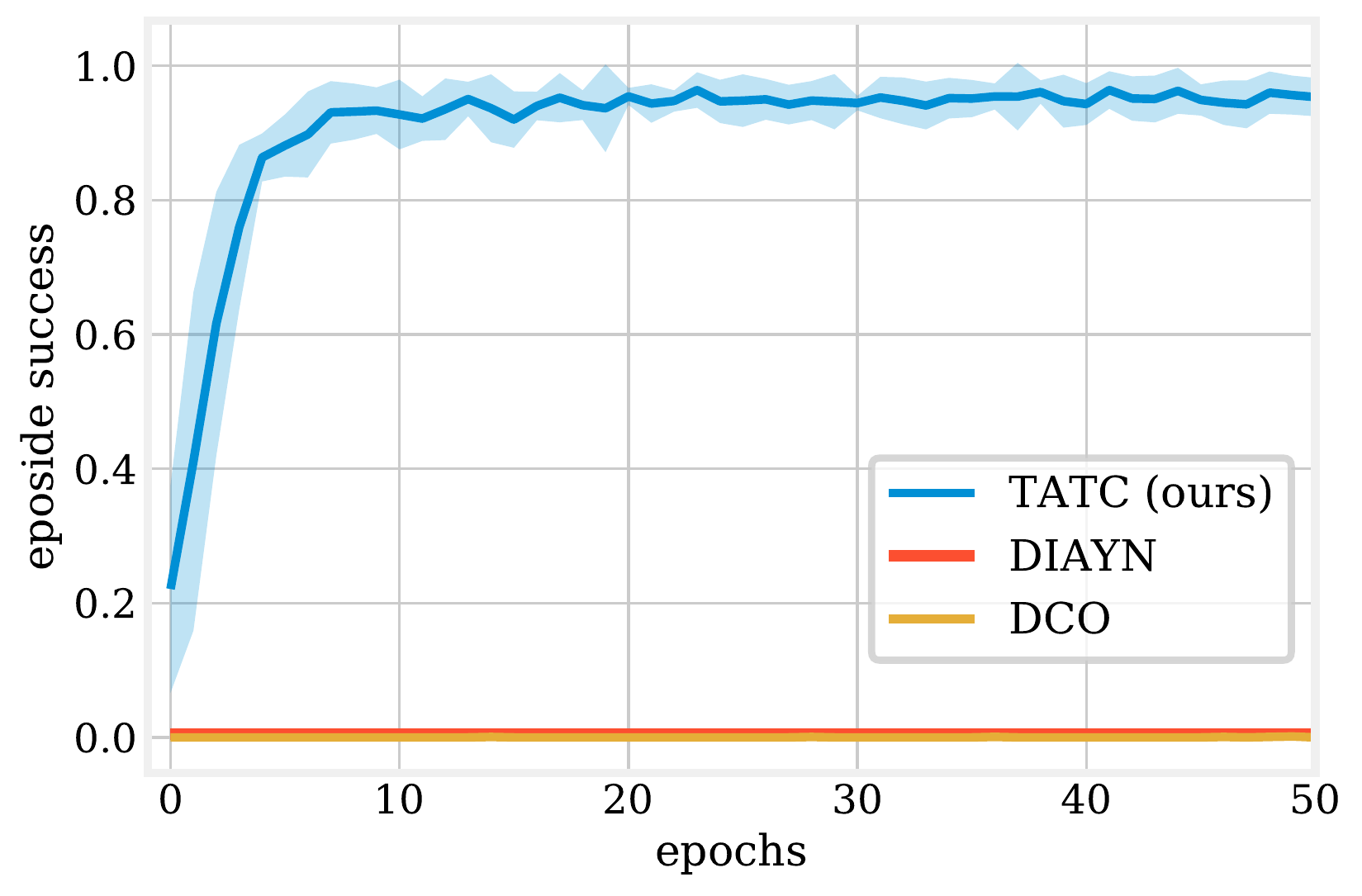}
    \caption{Skills Evaluation: performance gathered from $5$ independent runs. $95\%$ confidence intervals are shaded. Curves were exponentially smoothed (0.9) for better visualization.}
    \label{fig:skills_sparse}
\end{figure}

The sparsity of the reward naturally poses a challenge as no additional signal can guide the agent towards the goal, unlike the evaluation setting of DCO and DIAYN by~\citet{Jinnai2020Exploration}. \cref{fig:skills_sparse} shows that the skills learned by \textsc{TATC} quickly assist to complete the task while the skills learned with DCO and DIAYN do not. DIAYN's limited performance in difficult sparse-reward navigation tasks was also confirmed by~\citet{kamienny2021direct}. These results suggest that in order to succeed, DCO and DIYAN skills may require a richer signal like the distance-based dense reward used by~\citet{Jinnai2020Exploration} to evaluate both of them -- and where they show similar performances.

\section{Related Work}

Our work is related to self-supervised learning~\citep{bromley1994signature, chopra2005learning}, which brought recent advances in representation learning~\citep{bachman2019learning,he2020momentum,chen2020simple,grill2020bootstrap,caron2020}. These techniques have naturally been adapted to RL, especially contrastive methods. While some of these benefited from visually contrasting observations~\citep{laskin2020curl,yarats2021reinforcement}, others leveraged temporal contrasts to learn useful representations~\citep{mazoure2020,stooke2021decoupling,li2021learning}, which fall closer to our work.

We designed our covering policy as a hierarchical agent. This has actually been the default setting to model temporally-extended actions~\citep{Sutton1999}. Our work shares the same motivation as \citet{vezhnevets2017feudal} for training skills to follow latent directions. Among the large body of work on skill discovery, the eigenoptions framework \citep{Machado2017} and its extensions \citep{Machado2018,machado2021temporal,Jinnai2020Exploration} are probably the closest to our skill training scheme. Eigenoptions also fit in the directional skills definition as they are trained to travel along the directions defined by the eigenvectors of the Laplacian. These vectors have a dimensionality of $|\mathcal{S}|$ which can be very large. To contrast, our directional skills are defined by an arbitrary and diverse set of directions in the learned representation space of adjustable dimensionality, which offers more tractability. \textsc{TATC} skills and eigenoptions also share an interesting connection to mutual information based intrinsic control methods~\citep{Hansen2020Fast}, which we discuss in \cref{app:connect_visr}. Indeed, a similar discussion can be adapted to eigenoptions' intrinsic reward function.

The exploration mechanism in \textsc{TATC} emerges from the interplay between the representation and the covering policy rewarding scheme. As discussed in \cref{section:aug_rep_L}, the boredom term prevents the high-level policy from focusing exploration resources on previously over-sampled skills. This penalization induces \emph{optimism} towards the remaining skills, in the hope of deviating from previously explored regions. \citet{zhang2021made} share a similar motivation and derive an exploration bonus that aims at directly maximizing this deviation in terms of the policy occupancy. Both methods are inspired from the \emph{optimism-in-the-face-of-uncertainty} principle.

The incremental discovery paradigm has been previously adopted, either for exploration \citep{ecoffet2021nature}, incremental skill discovery \citep{Jinnai2020Exploration, pong2019skew}, or even state abstraction~\citep{misra2020kinematic}. Finally, we use learned skills to penalize boredom~\citep{schmid1991,oudeyer2009intrinsic} in the representation space and encourage exploration. The idea of using skills to foster curiosity has also been investigated by \citet{bougie2020skill}.

\section{Conclusion}
The Laplacian representation as proposed by~\citet{Wu2019} made the benefits of spectral methods affordable in large state spaces.
Unfortunately, the quality of this representation is strongly tied to the uniformity of its training data distribution, as shown is Section~\ref{experiments}. This has motivated the method proposed in this work where we reconcile a similar temporally-contrastive representation with exploration demanding settings. Our approach leverages the
practical skill training framework that such representations allow. The learned skills are used to better cover the state space and hence learn a better representation.
We validate our method in tabular as well as continuous environments.
Our representation learned in a non-uniform-prior setting shows 
a comparable representational power to the one acquired by a Laplacian representation from a uniform prior, and our skills proved to be competitive in hard continuous control tasks. With these results, we hope to bring such representations’ applicability one step closer to realistic contexts.

We have proposed to augment the representation objective with temporal abstractions captured in the acquired skills. This benefits exploration by inducing a boredom-fighting mechanism, and enforces the representation's dynamics-awareness. Intuitively, this augmentation can be seen as bringing temporally close regions (connected through skills) closer in the representation space. This observation may motivate further investigations on how temporal abstractions can improve the representational potential and benefit task-agnostic representations in RL.

\section*{Acknowledgement}
AE would like to thank Emmanuel Bengio, Amy Zhang, and Ahmed Touati for the insightful discussions, and Jad Kabbara and Riashat Islam for their feedback on an earlier version of this paper.
The authors acknowledge the financial support of CIFAR, NSERC and Mila.

\bibliography{main_tatc}

\clearpage

\appendix
\section{\underline{T}emporal \underline{a}bstractions augmented \underline{t}emporally-\underline{c}ontrastive learning (TATC) in the non-uniform prior setting}
\label{app:algo}

The proposed approach consists in simultaneously training the representation $\phi$ and the hierarchical agent $(\pi_\low, \pi_\hi)$.
The idea is to progressively extend the explored area while maintaining the previously collected knowledge.
To do so, in the non-uniform-prior setting, the agent switches with some probability $p_{rw}$ between following a uniformly random policy $\pi_\mu$ and executing the hierarchical policy (skills). The latter helps reach further areas more efficiently, where data collected by $\pi_\mu$ is used to train the representation $\phi$. Along their training, the skills progressively extend to cover newly discovered areas. Algorithm \ref{alg:simultaneous_training} provides a pseudocode of the proposed approach, in the non-uniform-prior setting.

\begin{algorithm}[h]
	\caption{\textsc{TATC} in the non-uniform prior setting}
	\begin{algorithmic}[1]
	    \STATE \textbf{Input:} $L, c, p_{rw}, N$
		\FOR {$iteration=1,2,\ldots$}
		    \STATE $D_{\pi_\mu}=\emptyset$, $D_s = \emptyset $
			\FOR {$batch=1,2,\ldots,N$}
			    \STATE Reset to $s_0$ with probability $p_r$.
			    \STATE $p \sim$ \emph{Unif}([0,1])
			    \IF{$p < p_{rw}$}
			        \STATE Run the uniformly random policy $\pi_\mu$ to collect $L$ random walk trajectories $\{\tau'_i\}_{i=1}^L$ of $c$ steps each.
			        \STATE $D_{\pi_\mu} \gets D_{\pi_\mu} \cup \{\tau'_k\}_{k=1}^L$
			     \ELSE
			        \STATE Run $(\pi_\hi,\pi_\low)$ to collect $L$ consecutive skills' trajectories $\{(\tau_k, \bm{\delta}_k)\}_{k=1}^L$ and their corresponding directions
			        \STATE $D_s \gets D_s \cup \{(\tau_k, \bm{\delta}_k)\}_{k=1}^L$
			    \ENDIF
			\ENDFOR
			\STATE Optimize the policies $(\pi_\hi,\pi_\low)$ using the intrinsic objectives \ref{eq:skill_reward} and \ref{eq:high_reward}  
			\STATE Optimize $\phi$ so as to minimize $\mathcal{L}_\text{TATC}(\phi; \mathcal{D}_s, \mathcal{D}_{\pi_\mu})$ (Equation ~\ref{eq:rep-L}).
		\ENDFOR
	\end{algorithmic}
	\label{alg:simultaneous_training}
\end{algorithm}

\section{Representation Objective Augmentation: Ablation Study}
\label{app:ablation}

\subsection{Boredom augmentation helps exploration}
In order to illustrate the importance of the boredom term $\mathcal{B}$ -- in the final objective (\ref{eq:rep-L}) we conducted the same representation learning experiments for the three gridworld domains in the non-uniform prior setting, but this time with the non-augmented representation learning objective ($\beta'=0$). 

\begin{figure}[ht!]
    \centering
    \begin{subfigure}[t]{0.2\columnwidth}
        \centering
        \includegraphics[width=\linewidth]{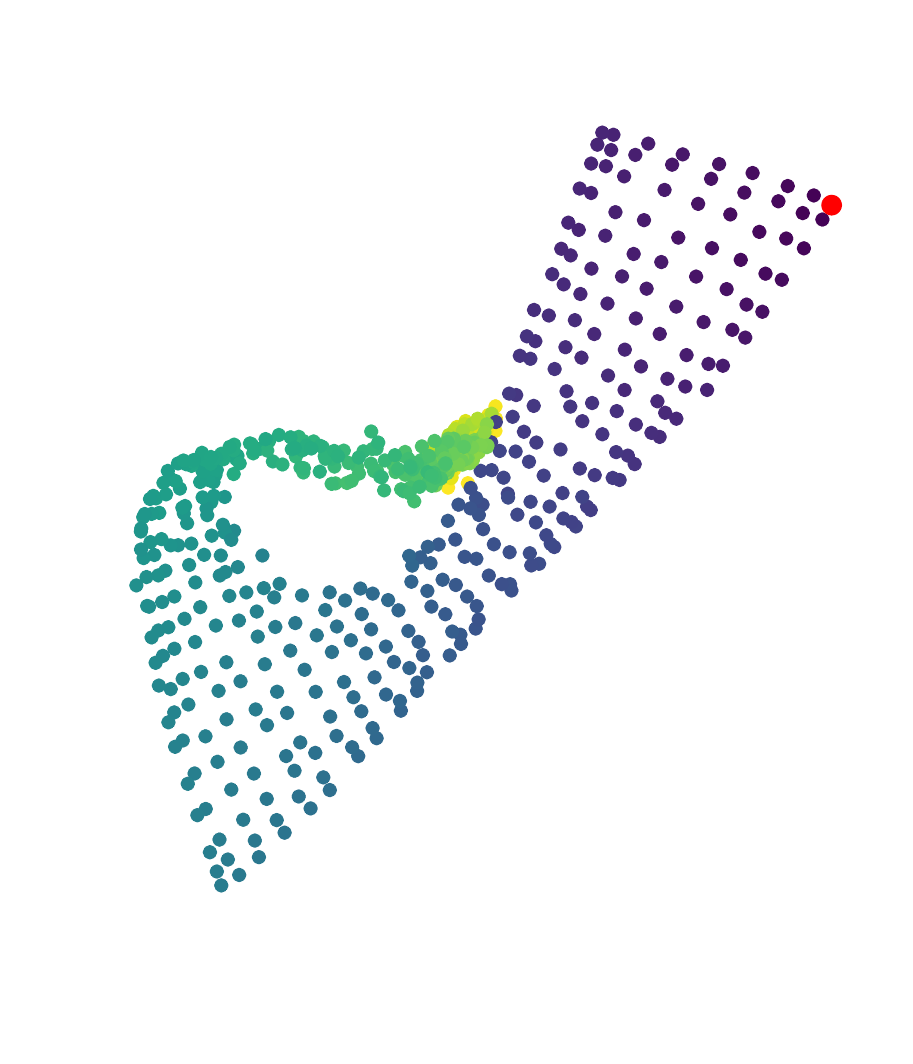}
        \caption{\textsc{U-Maze}}
    \end{subfigure}%
    ~ \hspace{0.1\columnwidth}
    \begin{subfigure}[t]{0.2\columnwidth}
        \centering
        \includegraphics[width=\linewidth]{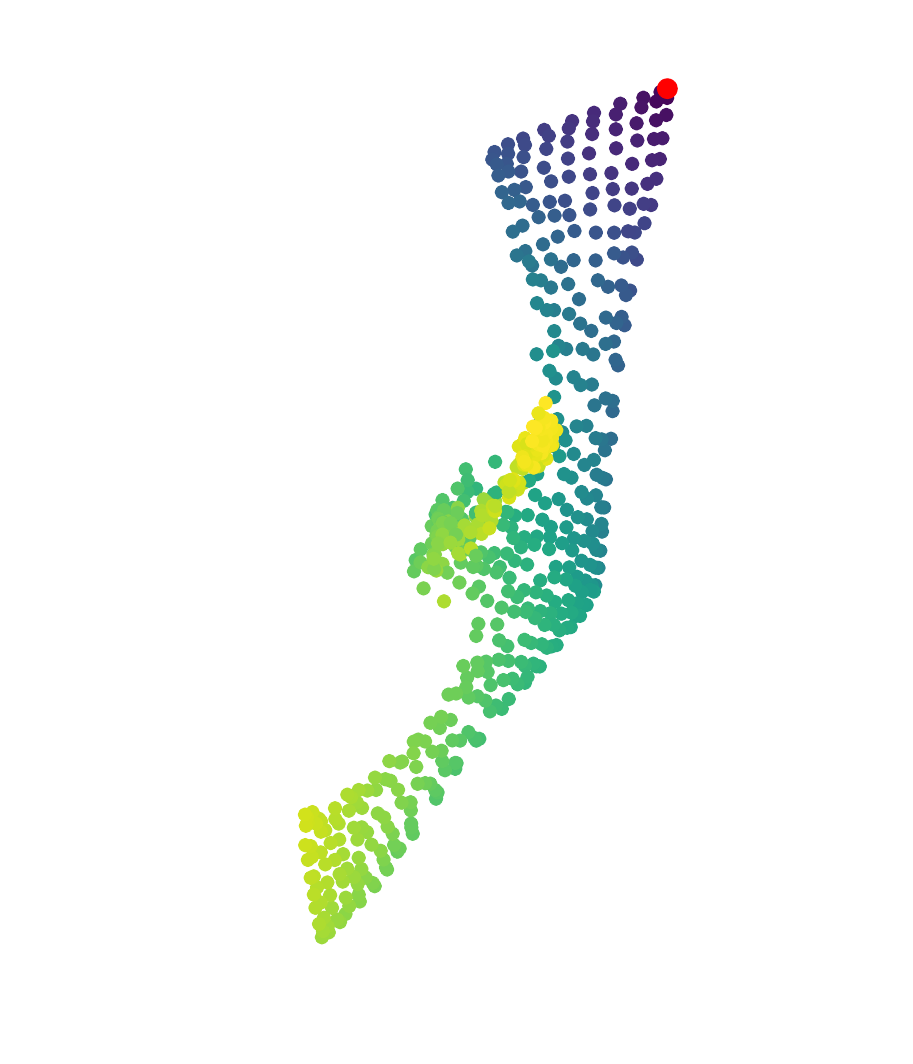}
        \caption{\textsc{T-Maze}}
    \end{subfigure}
    ~ \hspace{0.1\columnwidth}
    \begin{subfigure}[t]{0.2\columnwidth}
        \centering
        \includegraphics[width=\linewidth]{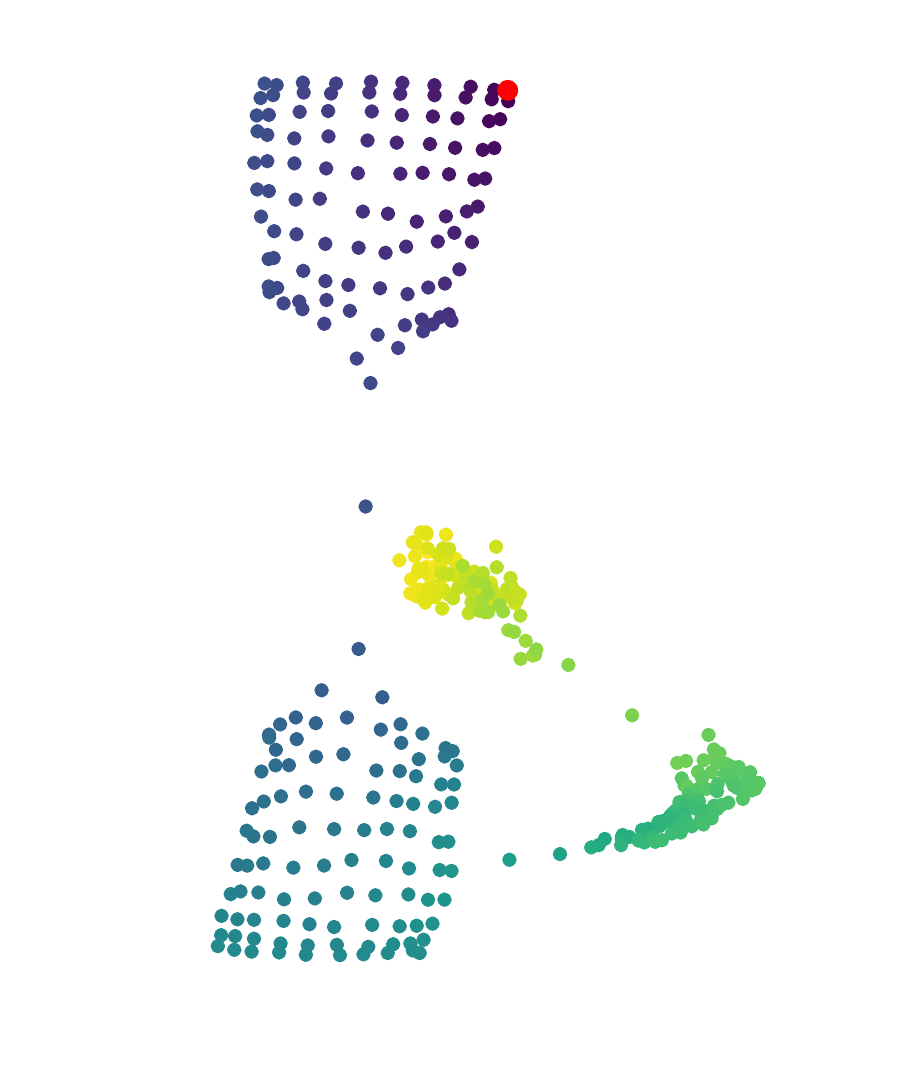}
        \caption{\textsc{4-rooms}}
    \end{subfigure}
    \caption{Learned representations in the gridworld domains with the \emph{non-augmented} objective. Without the boredom term, the agent fails to cover the state space (cf. Figure \ref{fig:progress_exp}), and may settle for incomplete representations. The colors reflect the distances in terms of the dynamics. They can be seen as quantities proportional to the length of the shortest path from the $s_0$ (marked in red) to the represented state.}
    \label{fig:ablation_exp}
\end{figure}

\cref{fig:ablation_exp} shows how the agent failed at exploring the whole domain. In \textsc{T-Maze}, it focuses only on one corridor without getting curious about the other one. Regarding \textsc{U-Maze} and \textsc{4-rooms}, the agent stops exploring after discovering the end of the first corridor and the second room, respectively. This is due to the lack of incentive to visit the yet unseen states, as they are less rewarding for $\pi_\hi$ (i.e., closer in the representation space, hence smaller $R^\hi$) than the furthest explored state. The effect of the proposed augmentation compresses the representation of the explored area, say the first corridor in \textsc{U-Maze}, which makes the rest of the environment more appealing to explore for $\pi_\hi$ (i.e. further in the representation space, hence larger $R^\hi$).

\subsection{Boredom augmentation enforces dynamics-awareness}
\label{app:dynamic_aware}
\begin{figure}[ht!]
    \centering
    \begin{subfigure}[t]{0.25\columnwidth}
        \centering
        \includegraphics[width=\linewidth]{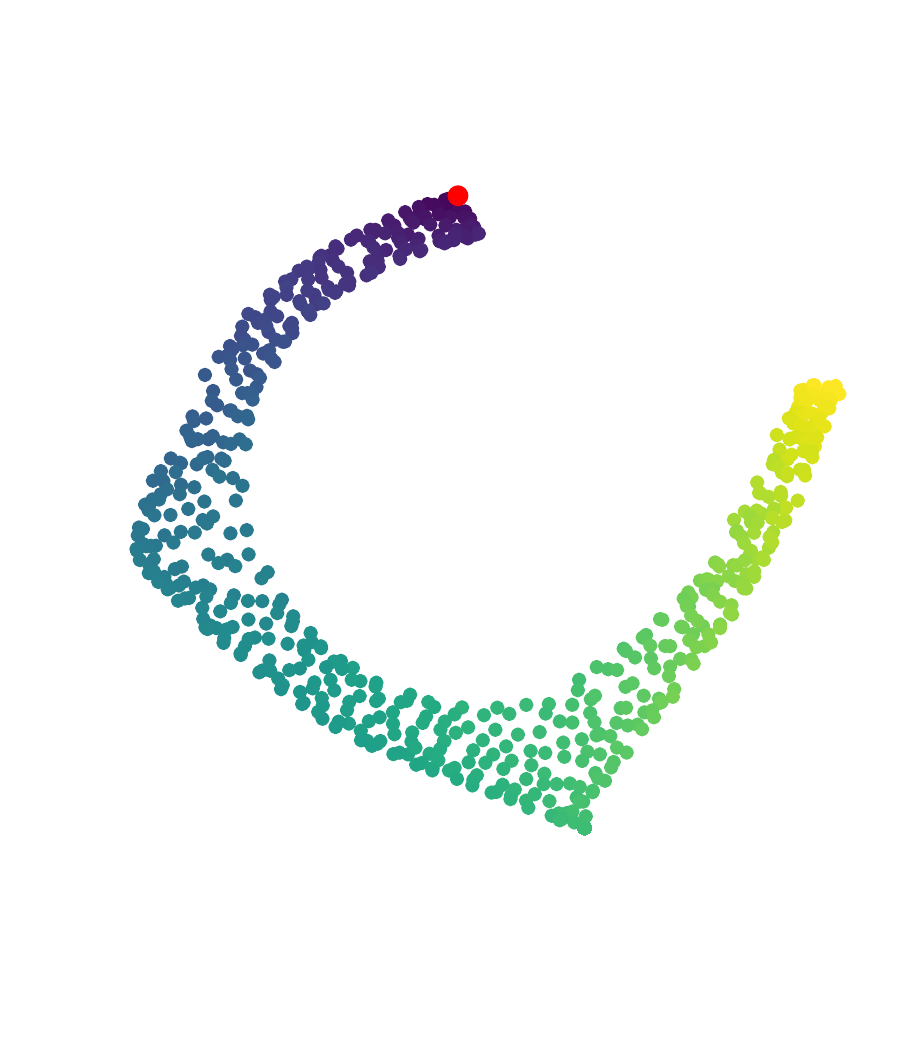}
        \caption{U-Maze}
    \end{subfigure}%
    ~ \hspace{0.1\columnwidth}
    \begin{subfigure}[t]{0.25\columnwidth}
        \centering
        \includegraphics[width=\linewidth]{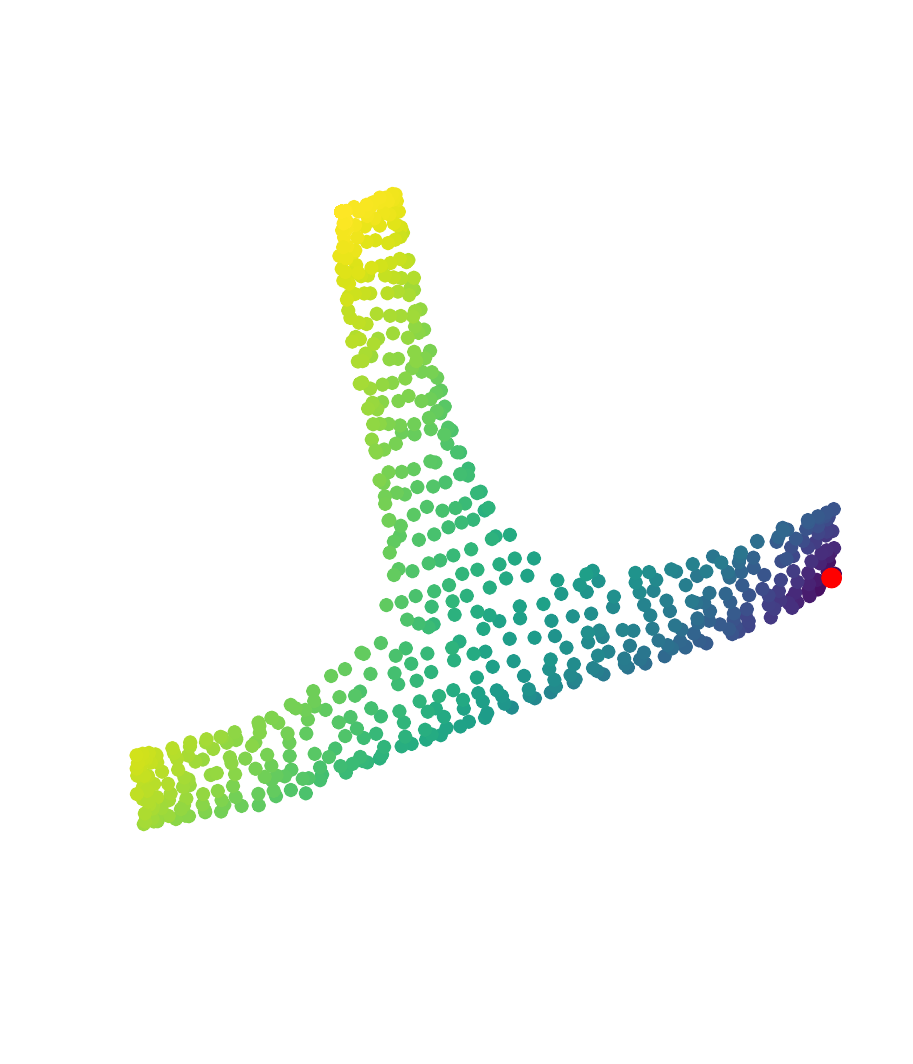}
        \caption{T-Maze}
    \end{subfigure}
    ~ \hspace{0.1\columnwidth}
    \begin{subfigure}[t]{0.25\columnwidth}
        \centering
        \includegraphics[width=\linewidth]{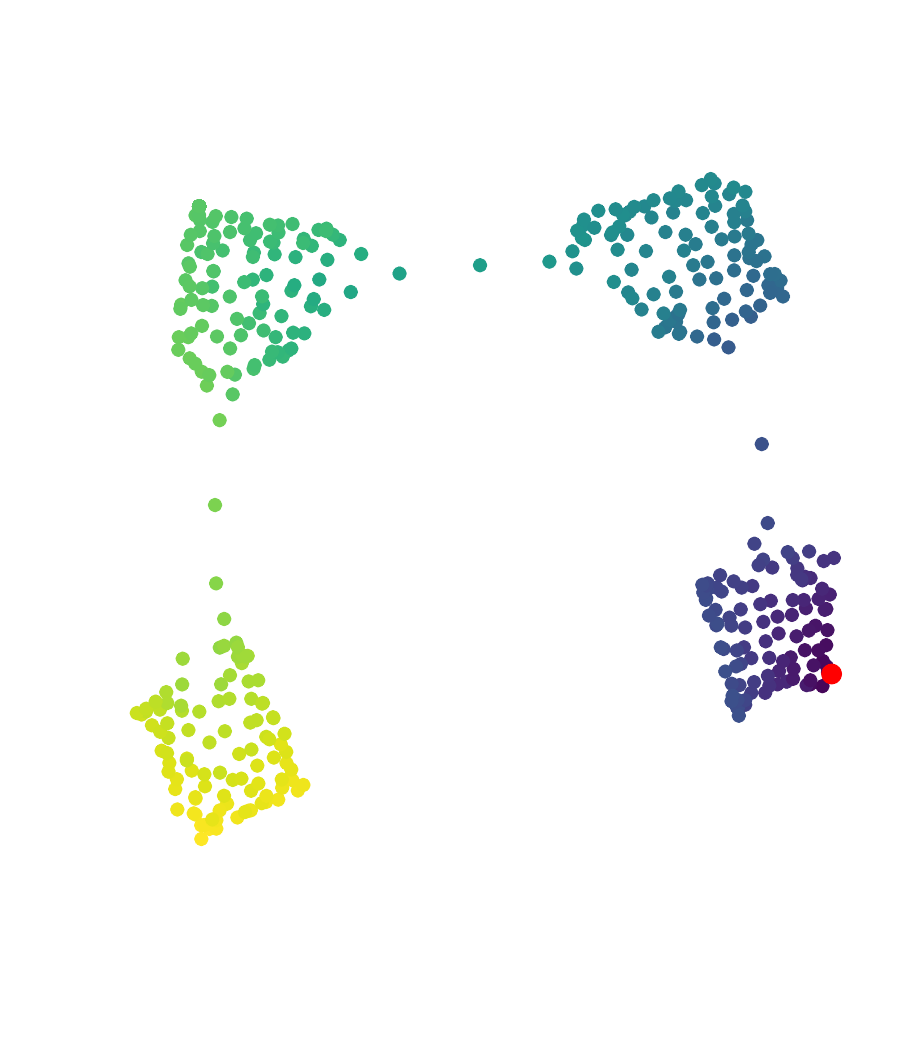}
        \caption{4-rooms}
    \end{subfigure}
    \caption{Learned representations when uniformly sampling over the state space. Without the boredom term, the representation does not reflect temporally-extended dynamics. The colors reflect the distances in terms of the dynamics. They can be seen as quantities proportional to the length of the shortest path from the $s_0$ (marked in red) to the represented state.}
    \label{fig:ablation_dynamics_awrns}
\end{figure}

To verify the benefit of the boredom term beyond helping exploration, we train the representation with the non-augmented objective ($\beta'=0$) but this time in the uniform prior setting, so that to marginalize the exploration problem. Figure \ref{fig:ablation_dynamics_awrns} illustrates the learned representations in the three gridworld domains.
These representations have failed to capture the dynamics. For example, in the case of \textsc{4-rooms}, the distances from the first room to the fourth and third rooms are comparable in the representation space, which indicates that the representation does not take into account the relative order in which the rooms should be visited, when moving from the first room to the last.
Similarly, in \textsc{U-Maze}, the end of the maze is closer to the initial area than the second corner is. However, in order to reach the former on must pass by the latter.
This proves that the boredom term is not only important for the desired exploratory behavior (cf. Figure~\ref{fig:ablation_exp}), but also enhances the dynamics-awareness of our representation.

\section{Implementation details}
\label{app:implem}

\subsection{GridWorld}
The states are one-hot encoded such that no positional information is provided to the agent. The domains dimensions are: \textsc{U-Maze} $30\times30$, \textsc{T-Maze} $40\times30$, \textsc{4-rooms} $21\times21$.

For all the experiments, we defined the representation network as an MLP of two hidden layers of size $128$ with tanh activations and a linear output layer of the size of representation's dimensionality $d$. The high-level and the low-level policies are both MLPs of two hidden layers of size $128$ with tanh activations and a logsoftmax output layer of the size of their respective action spaces: the environment's $4$ actions for the low-level policy and $8$ actions for the high-level policy corresponding to the $8$ directions $\Omega = \{(\cos(2k\pi/n), \sin(2k\pi/n))\ |~ k \in \{0,...,7\}\}$ that define diverse skills.

The policies were trained with vanilla A2C with MC returns from the collected trajectories (Monte-Carlo estimates), i.e. no bootstrapped values where used. The skills being of a fixed size they could be trained without any reward discount ($\gamma=1$). 
The high-level and low-level policies were entropy-regularized with coefficients $0.3$ and $0.1$ respectively.

All of these networks were trained with RMSprop~\citep{hinton2012neural} and a step size of $0.001$. Environments specific hyperparameters are provided below.

\subsubsection{Representation Learning}

\paragraph{\textsc{U-Maze}.} Our representation is learned in the non-uniform prior setting with $p_r{=}0.3$, $p_{rw}{=}0.4$ and $K{=}90$ (around the number of steps between $s_0$ and the furthest state in the maze).
We learn a 2-dimensional representation ($d=2$) using the representation learning objective \ref{eq:rep-L} with $\beta=0.2$ and $\beta'=2$.
We fix the skills length to $c=30$ steps (so $L=K/c=3$), and jointly train the representation $\phi$ and the policies $(\pi_\hi,\pi_\low)$ by collecting, for each update, a batch of $N=32$ trajectories of length $c$ to fill $D_s$ and $D_{\pi_\mu}$ as described in Algorithm~\ref{alg:simultaneous_training}. We train them for $700$ epochs where each epoch corresponds to $10$ updates (convergence to the complete representation required around $500$ epochs).

\paragraph{\textsc{T-Maze}.} Our representation is learned in the non-uniform prior setting with $p_r{=}0.2$, $p_{rw}{=}0.4$ and $K{=}40$ (around the number of steps between $s_0$ and the furthest state in the maze).
We learn a 2-dimensional representation ($d=2$) using the representation learning objective \ref{eq:rep-L} with $\beta=0.2$ and $\beta'=2$.
We fix the skills' length to $c=20$ steps (so $L=K/c=2$).
and jointly train the representation $\phi$ and the policies $(\pi_\hi,\pi_\low)$ by collecting, for each update, a batch of $N=32$ trajectories of length $c$ to fill $D_s$ and $D_{\pi_\mu}$ as described in Algorithm~\ref{alg:simultaneous_training}. We train them for $700$ epochs where each epoch corresponds to $10$ updates (convergence to the complete representation required around $350$ epochs).

\paragraph{\textsc{4-rooms}.} Our representation is learned in the non-uniform prior setting with $p_r{=}0.25$, $p_{rw}{=}0.5$ and $K{=}60$ (around the number of steps between $s_0$ and the furthest state in the maze).
We learn a 2-dimensional representation ($d=2$) using the representation learning objective \ref{eq:rep-L} with $\beta=0.2$ and $\beta'=2$.
We fix the skills' length to $c=20$ steps (so $L=K/c=3$).
and jointly train the representation $\phi$ and the policies $(\pi_\hi,\pi_\low)$ by collecting, for each update, a batch of $N=32$ trajectories of length $c$ to fill $D_s$ and $D_{\pi_\mu}$ as described in Algorithm~\ref{alg:simultaneous_training}. We train them for $700$ epochs where each epoch corresponds to $10$ updates (convergence to the complete representation required around $350$ epochs).

\paragraph{The Laplacian representation} (\textsc{Lap-rep}) was trained in the same environments' settings described above, for both the uniform and non-uniform prior settings (of course no policy is trained here so $p_{rw}=1$, and $(s_0, p_r)$ are not relevant for the uniform prior setting). Besides the representation's dimension $d$, we used the training configuration and hyperparameters proposed by~\citet{Wu2019}. For the uniform prior setting, our online data collection does not cause any discrepancy compared to the offline scheme used in~\citet{Wu2019}. Indeed, for a minibatch size large enough, the stochastic minibatch based training of \textsc{Lap-rep} when using a uniform prior is agnostic to the data collection scheme (offline vs online) since in both cases the minibatches are sampled from the exact same state distribution.

\subsubsection{Prediction and Control}

In the prediction and control experiments, we evaluate each pretrained representation by training an actor-critic agent to solve a goal-achieving task with a sparse reward ($r=1$ upon reaching the goal, and $r=0$ otherwise). Here are the set goal positions: $(1,30)$ in \textsc{U-Maze}, $(25,30)$ in \textsc{T-Maze} and $(1,21)$ in \textsc{4-rooms}.
The episode size was set to $100$ steps for all the gridworld domains.

For the prediction, the critic head is a linear function in the given representation, while the actor is an MLP with two hidden layers of size $64$ and tanh activations, a logsoftmax output layer of size $4$ (discrete gridworld actions), and the actor's input is the state one-hot code. For the control experiments, the actor-critic agent is defined on top of the representation as an MLP of two hidden layers of size $64$ with tanh activations that feed two output heads: a linear critic head, and a logsoftmax action head for the $4$ actions.
The agent is trained with A2C with MC returns and a discount of $\gamma=0.98$, a batchsize of $80$ episodes, an entropy regularization with a $0.01$ coefficient, and Adam optimizer~\citep{kingma2014adam} with a learning rate of $0.001$.
\subsection{MuJoCo: AntMaze}

The Ant agent has a 29 dimensional state space and a 8 dimensional action space (4 legs with 2 joints each to control). For the sake of simplifying the RL training algorithm,
we mapped each action-dimension interval to a discrete set of 5 values equally spaced over this interval.

\begin{figure}[H]
\centering
\begin{subfigure}[t]{0.35\columnwidth}
    \centering
    \includegraphics[width=\linewidth]{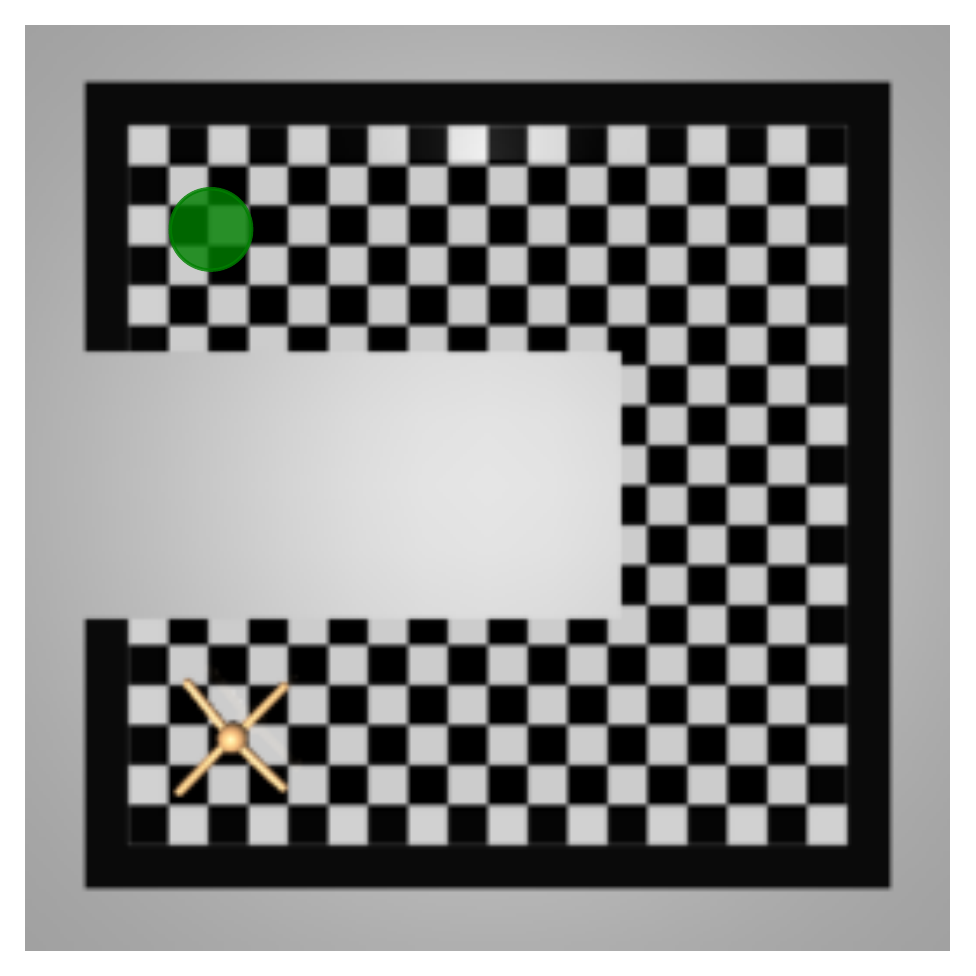}
    \caption{\textsc{AntMaze-1}}
\end{subfigure}
\hspace{0.05\columnwidth}
\begin{subfigure}[t]{0.35\columnwidth}
    \centering
    \includegraphics[width=\linewidth]{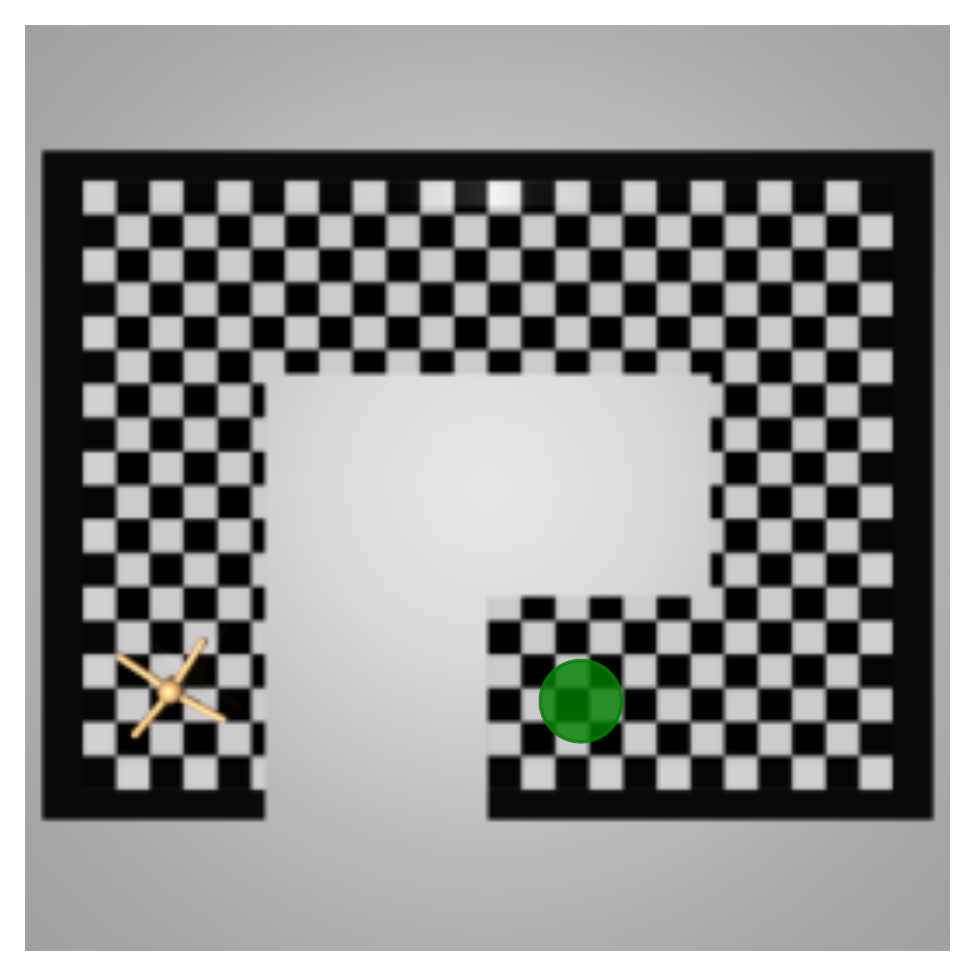}
    \caption{\textsc{AntMaze-2}}
\end{subfigure}%
\caption{\textsc{Antmaze} domains. Goal positions of the evaluation tasks are shown in green.}
\label{fig:antmaze_domains}
\end{figure}

We used 2 mazes similar in shape to those from \citet{Wu2019} (see \cref{fig:antmaze_domains}): \textsc{AntMaze-1} defines a 3D U-shaped corridor and \textsc{AntMaze-2} is a 3D swirl-shaped corridor.

\vspace{-3ex}

We used the same architectures for the representation and the policies as for the gridworld, with the only difference that for the low-level policy, the action head was adapted to the discretization of the action space by having $8$ logsoftmax output heads of size $5$, one for each action dimension, and the corresponding $5$ discrete values. This choice makes the training algorithm simpler as it allows using A2C here as well. 

Our representation is learned in the non-uniform prior setting with  $p_r=0.2$, $p_{rw}=0.3$ and $K=500$. We learn a 2-dimensional representation ($d=2$) using the representation learning objective \ref{eq:rep-L} with $\beta=0.2$ and $\beta'=5$.
We fixed their length to $c=100$ steps (so $L=K/c=5$).
and jointly train the representation $\phi$ and the policies $(\pi_\hi,\pi_\low)$ by collecting, for each update, a batch of $N=32$ trajectories of length $c$ to fill $D_s$ and $D_{\pi_\mu}$, as described in Algorithm~\ref{alg:simultaneous_training}. We train them for $1000$ epochs where each epoch corresponds to $10$ updates (convergence to the complete representation required around $650$ epochs).

The policies were trained with the same A2C used in gridworld domains and the same RMSprop hyperparameters. The high-level and low-level policies were entropy-regularized with the coefficients $0.15$ and $0.1$, respectively.

For the reward shaping and skills evaluation experiments, the goal positions are shown in \cref{fig:antmaze_domains}, and success is defined as being within $\epsilon$ from the goal. Here, $\epsilon=2$ which corresponds to half the size of the building blocks of the mazes.

\subsubsection{Reward Shaping}

For these experiments, \textsc{Lap-rep} was trained in both the uniform and the non-uniform prior setting. 

For the uniform prior setting, we used $d=2$ and followed the experimental framework of \citet{Wu2019}. Since our \textsc{AntMaze} environments are larger, we collected $500,000$ training samples (10 times more than in \citet{Wu2019}) from a uniformly random policy, then we trained the representation on this large dataset. For all other hyperparameters, we used those provided in \cite{Wu2019}. With $d=20$, our replication of \textsc{Lap-rep} did not succeed in reward shaping.

Regarding the non-uniform prior setting, we used the same setting configuration as for \textsc{TATC}, with the representation objective from~\citet{Wu2019}.
We have tested online (similar to \textsc{TATC}) and offline \citep{Wu2019} data collection for the representation training, and $d\in \{2, 20\}$. Both schemes ended up performed the same way for the reward shaping task.

Now, for the reward shaping, we train a Soft Actor-Critic (SAC)~\citep{haarnoja2018soft} agent to reach a goal area (neighbourhood around the goal state) with episodes of size $1000$ steps. We use the following hyperparameters: 
\begin{itemize}
    \item Discount $\gamma=0.99$
    \item Entropy coefficient (temperature) $\alpha=0.1$
    \item Soft critic updates with smoothing constant $\tau=0.005$
    \item Replay buffer of size $5\cdot10^6$ (equal to the number of training steps).
    \item Adam optimizer with step size of $0.0001$
\end{itemize}

As SAC is sensitive to the reward scale~\citep{haarnoja2018soft}, we grid-searched this hyperparameter in $\{10^{-5}, 10^{-4}, \cdots, 1, 2, 10, 20\}$, and the best performing one for our representation was $1$ in \textsc{AntMaze-1} and $0.01$ in \textsc{AntMaze-2}.
Regarding \textsc{Lap-rep}, we found that $10$ and $0.01$ worked the best for these two mazes, respectively. All these coefficients correspond to the dense reward shaping setting. Their values were doubled for the half-half mix reward setting, to account for the $0.5$ coefficient.

\subsubsection{Skills Evaluation}
\label{app:skill_eval}

To train DCO, we first collect a dataset to estimate the second eigenvector and then use the same dataset to train a policy -- the option -- using DDPG~\citep{lillicrap2015continuous}. Each DCO option is tied to its own eigenvector estimate and its own training set of size $500000$ (10 times the size used in~\citet{Jinnai2020Exploration}). As suggested by the authors of DCO \citep{Jinnai2020Exploration}, the remaining hyperparameters to estimate the eigenvectors and train their corresponding options were taken from \citet{Wu2019}. DIAYN skills were trained as recommended by \citet{Eysenbach19}. For fair comparison, we train $8$ skills for both DCO and DIAYN.

For the skills evaluation stage, we freeze the learned low-level policies and train a high-level policy to use the $8$ skills as the only available actions to reach the goal $g$ on the other end of the \textsc{AntMaze-1} environment using a sparse reward $r_t = \mathds{1}\left [\|s_{t+1}-g\|_2\leq \epsilon \right ]$ within a finite horizon of $1000$ steps. Note that this tasks is quite challenging given the type of reward and the length of episode especially in a continuous state space. As our skills offer some flexibility in their execution (can be started everywhere and run for arbitrary number of steps), this episode length was decomposed to $5$ skills of $200$ steps each. The high-level policy was trained with A2C with MC returns (no discount) a batch size of $8$ episodes, and RMSprop optimizer with a learning rate of $0.001$.

\section{The switching utility of the boredom term}
Note that $\mathcal{D}_s$, in \cref{eq:boredom}, may contain trajectories from skills that are not yet duly trained; for example early in the training or in a freshly discovered area. Since at that stage, these skills' trajectories are close to random walks, their contribution in the boredom term is similar to the first attractive term, in \cref{eq:base_obj_empir}. This means that a new skill trajectory initially contributes to the temporal similarity term (attractive term) in training the representation, thus making the most out of the sampled skills' trajectories while these are still early in their training. 
The more a skill is trained, the more structured its trajectories become and the more they contribute to the intended "boredom" effect (\cref{section:aug_rep_L}), that is encouraging exploration and dynamics awareness (Appendix~\ref{app:ablation}). 

\section{Connection to Behavioral Mutual Information}
\label{app:connect_visr}

There are numerous methods in the \emph{intrinsic control} literature that aim at maximizing the mutual information between the agent's behavior and a conditioning variable that encodes the available skills.

This type of intrinsic control is achieved by training a skill conditioned policy, $\pi$, to maximize the mutual information between the skill code, $z$, and some representation of the trajectory, $\tau$, obtained from the conditioned policy $\pi(\cdot|z)$. This objective can be written as:
\begin{equation}
\label{eq:MI}
    I(z; f(\tau)) = \mathcal{H}(z) - \mathcal{H}(z| f(\tau)),
\end{equation}
where entropy is denoted by $\mathcal{H}$, and $f$ is a function of the trajectory. It is also common to assume that $z$ is sampled from a fixed prior~\citep{Eysenbach19}, which simplifies this policy training objective as a minimization of the conditional entropy term in \cref{eq:MI}. In practice, the adopted training loss, can be derived as a lower bound of this quantity, using an approximate posterior, $q$:
\begin{equation}
    L_q(\pi) = - \mathbb{E}_{z,\pi}[\log q(z| f(\tau)) ].
\end{equation}
Traditionally, the integrand of this expectation defines the intrinsic reward of the skill conditioned policy, while the approximate posterior $q$ is trained to discriminate the correct $z$ based on the observed behavior $\tau$. 

Inspired by the fast inference offered by Successor Features, \citet{Hansen2020Fast} proposed to use a \emph{log-linear} discriminator in the successor representation (SR), $\phi_\text{SR}$. In other words, the skill rewards can be written as:
\begin{equation}
\label{eq:visr_rew}
    r(s) = \phi_\text{SR}(s)^\top \mathbf{w},
\end{equation}
with $\mathbf{w}$ playing the role of the skill-identifying variable (denoted by $z$ in the general case above). Note that in this case, the function $f$ maps, as it is commonly the case, to the final state of the trajectory. Now, consider the case where the trajectory is instead represented by the (normalized) latent direction of the final transition $(s,s')$. This reward would be
\begin{equation}
\label{eq:visr_rew_2final}
r(s,s') = \frac{(\phi_\text{SR}(s')-\phi_\text{SR}(s))^\top \mathbf{w}}{\|\phi_\text{SR}(s')-\phi_\text{SR}(s)\|_2}.
\end{equation}
Let's recall that the SR\footnote{The SR is encoded by the matrix $(I-\gamma T)^{-1}$, with $T$ the MDP's transition matrix.} shares the same eigenvectors as the Laplacian~\citep{stanchenfeld14, Machado2018}. This implies that it can also be approximated with a temporally-contrastive objective~\citep{Wu2019}, and potentially replaced by our alternative \textsc{TATC} representation $\phi$. Finally, we can rewrite \cref{eq:visr_rew_2final} as
\begin{equation}
\label{eq:tatc_visr}
r(s,s') = \frac{(\phi(s')-\phi(s))^\top \mathbf{w}}{\|\phi(s')-\phi(s)\|_2}
\end{equation}
This reward corresponds to our skills intrinsic reward from \cref{eq:skill_reward}, with $\mathbf{w} \equiv \bm{\delta}$.

\end{document}